\documentclass[]{fairmeta_gr2}

\usepackage{multirow}
\usepackage{color}
\usepackage{colortbl}
\usepackage{mdframed}
\usepackage{caption}
\usepackage{subcaption}
\usepackage{algorithm}
\usepackage{algpseudocode}
\usepackage{amsthm}
\newtheorem{definition}{Definition}
\newtheorem{oracle}{Oracle}
\usepackage{amssymb}
\usepackage{wrapfig}
\usepackage{adjustbox}
\usepackage{makecell}
\usepackage{lmodern}

\usepackage{enumitem}

\title{A Replicate-and-Quantize Strategy for Plug-and-Play Load Balancing of Sparse Mixture-of-Experts LLMs}

\author[1]{Zijie~Liu}
\author[1]{Jie Peng}
\author[1]{Jinhao Duan}
\author[2]{Zirui Liu}
\author[3]{Kaixiong Zhou}
\author[4]{Mingfu Liang}
\author[4]{Luke Simon}
\author[4, \dagger]{Xi Liu}
\author[5, \dagger]{Zhaozhuo Xu}
\author[1, \dagger]{Tianlong Chen}

\affiliation[1]{University of North Carolina at Chapel Hill}
\affiliation[2]{University of Minnesota Twin Cities}
\affiliation[3]{North Carolina State University}
\affiliation[4]{Meta AI}
\affiliation[5]{Stevens Institute of Technology}

\contribution[\dagger]{Joint correspondence author}

\abstract{Sparse Mixture-of-Experts (SMoE) architectures are increasingly adopted to scale large language models (LLMs) efficiently, delivering strong accuracy under fixed compute budgets. However, SMoE architectures often suffer from severe load imbalance across experts, whereby a minority of experts are dispatched the majority of tokens while the remainder are underutilized. Prior research has largely focused on improving training-time load balancing through routing regularization or auxiliary losses, leaving the inference-time behavior—crucial for practical deployment—relatively underexplored.
To bridge this gap, we conduct a systematic analysis of expert routing behavior during the inference stage and uncover three key insights: (i) load imbalance persists at inference time and intensifies as batch size increases, (ii) selection frequency does not reliably indicate expert importance and (iii) the overall workload and importance of all experts can be estimated from a small calibration set. These findings motivate the design of inference-time mechanisms that dynamically balance expert workloads without retraining or modifying the router.
In this paper, we introduce the Replicate-and-Quantize (R\&Q) strategy, a training-free and near-lossless framework for dynamically rebalancing expert workloads during inference. The core idea is, in each layer, to replicate heavy-hitter experts, which receive the most tokens, to provide additional parallel capacity; meanwhile, less important experts, which contribute less to overall performance, are quantized together with the replicas to remain within the original memory budget. Moreover, to quantify imbalance, we further propose a Load-Imbalance Score (LIS) computed per MoE layer, which compares the number of tokens assigned to the heavy-hitter expert with the number each active expert would receive under an equal split. LIS equals one for perfectly balanced routing and increases proportionally as imbalance grows. Extensive experiments across representative SMoE architectures and benchmarks demonstrate that R\&Q achieves up to 1.4× reduction in load imbalance while maintaining accuracy within ±0.6\%, offering a practical path toward predictable and efficient SMoE inference.}

\date{\today}

\begin{document}

\maketitle

\section{Introduction}
Meta authors were not involved in conducting the experiments.\footnote{Meta authors were not involved in conducting the experiments.}

The rapid scaling of large language models (LLMs) has driven unprecedented advances in natural language processing (NLP). 
Empirical studies show that model quality improves systematically as parameter count, dataset size, and compute increase, 
a trend formalized as scaling laws~\cite{kaplan2020scalinglawsneurallanguage}. 
These scaling behaviors have motivated the construction of frontier LLMs such as GPT-3 (175B parameters) and PaLM (540B parameters)~\cite{chowdhery2022palmscalinglanguagemodeling}, 
which exhibit strong generalization across domains including mathematical reasoning~\cite{wang2023mathcoder, imani2023mathprompter}, 
code generation~\cite{yuan2023scaling}, and multi-hop question answering~\cite{huang2022towards}. 
However, these models contain billions of parameters, requiring immense computational resources for both training and inference~\cite{chowdhery2022palmscalinglanguagemodeling}. 
In practice, this leads to a pronounced efficiency gap: While training can be carried out in large-scale datacenters with abundant computational resources, model deployment typically takes place in latency-sensitive and resource-limited settings such as online services or edge platforms, where comparable compute and memory capacity are unavailable.
As a result, inference efficiency-rather than training efficiency-has become the primary obstacle to real-world deployment, 
where latency, throughput, and cost directly determine usability.

To mitigate these issues, recent work has emphasized architectural efficiency, aiming to design model structures that maintain high representational capacity without proportionally increasing the computation required per token. Among various scaling techniques, one of the most effective is the \textit{Sparse Mixture-of-Experts} (SMoE) architecture~\cite{shazeer2017outrageously, fedus2022switch,chen2023sparse, zhao2023sparse}. 
In SMoEs, only a small subset of experts is activated for each input token based on routing scores, 
allowing the model to scale its capacity without linearly increasing computation. 
This sparse activation allows different experts to specialize in particular subsets of inputs, so that computation is allocated selectively rather than uniformly across tokens.
SMoE designs have been widely adopted in state-of-the-art systems such as Switch Transformer~\cite{fedus2022switch}, DeepSeek-MoE~\cite{dai2024deepseekmoe}, and Mixtral~\cite{jiang2024mixtralexperts}, 
making SMoE a mainstream paradigm for scaling LLMs efficiently during both training and inference.

\begin{wrapfigure}{r}{0.48\columnwidth}
\vspace{-10pt}
\centering

\includegraphics[width=0.48\columnwidth]
{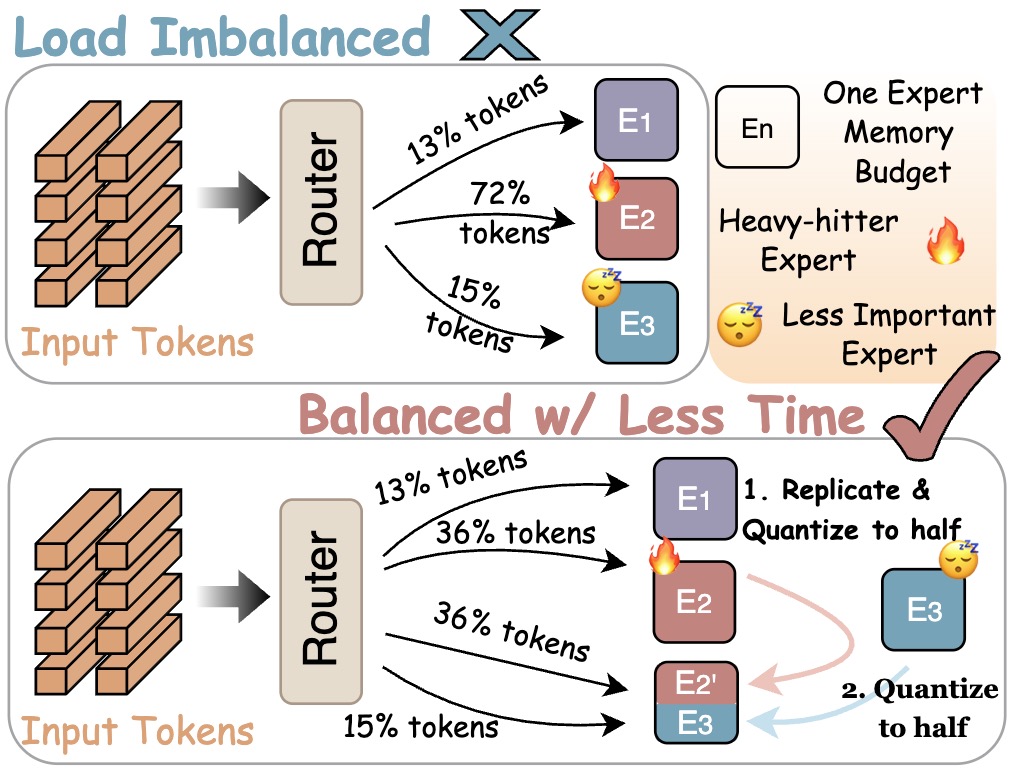}

\caption{
\textbf{Overview of the proposed Replicate-and-Quantize (R\&Q) framework.}
Compared to vanilla SMoE inference, which suffers from severe load imbalance due to overused heavy-hitter experts,
R\&Q mitigates this issue by replicating high-load experts with quantized copies while simultaneously quantizing less important ones within the same memory budget.
This achieves more balanced expert utilization and faster inference without requiring model retraining.
}

\label{fig:moe_small}
\vspace{-2pt}
\end{wrapfigure}

\textbf{Challenges in SMoE Inference.}  
Despite their widespread success, Sparse Mixture-of-Experts (SMoE) models still suffer from persistent load imbalance across experts during both training and inference. Load imbalance refers to the uneven distribution of tokens among experts in an SMoE layer, where certain experts—referred to as heavy-hitter experts—are disproportionately selected by the routing mechanism, while others receive few or no tokens. In practice, the resulting token-to-expert assignments are often heavy-tailed: a small subset of experts receives most routed tokens, while many others are rarely activated. This workload skew leads to significant inefficiency during inference, as overloaded experts become bottlenecks and underused experts waste computational resources. First, heavy-hitter experts become stragglers that determine the batch completion time, thereby increasing end-to-end latency~\cite{he2025capacityawareinferencemitigatingstraggler, doucet2025harmoenyefficientmultigpuinference}. Second, experts that receive few or no tokens leave compute resources idle and depress effective parallelism, yielding suboptimal device utilization~\cite{Dhasade_2025, he2025capacityawareinferencemitigatingstraggler}. Third, persistent workload skew increases synchronization and buffer-management overhead during routing and dispatch. When accumulated at scale, these overheads lead to higher energy consumption and operational costs.~\cite{he2025capacityawareinferencemitigatingstraggler, doucet2025harmoenyefficientmultigpuinference}. Without mitigation, these issues can substantially reduce the efficiency benefits that SMoEs are designed to achieve.

Thus, many previous studies have focused on mitigating load imbalance during the training phase through techniques such as auxiliary losses~\cite{lepikhin2020gshard}, which explicitly penalize uneven expert utilization to encourage uniform token distribution, entropy regularization~\cite{zhou2022mixture}, which increases routing diversity and prevents router collapse, and reinforcement-based routing policies~\cite{clark2022unifiedscalinglawsrouted}, which learn adaptive expert selection through reward signals that balance accuracy and utilization.
These methods explicitly regularize the routing behavior to promote more uniform token-to-expert assignments and prevent expert underutilization.
As a result, they effectively improve load balance and enhance training stability within the training distribution.

\begin{wrapfigure}{l}{0.5\textwidth}
    \centering
    \includegraphics[width=0.48\textwidth]{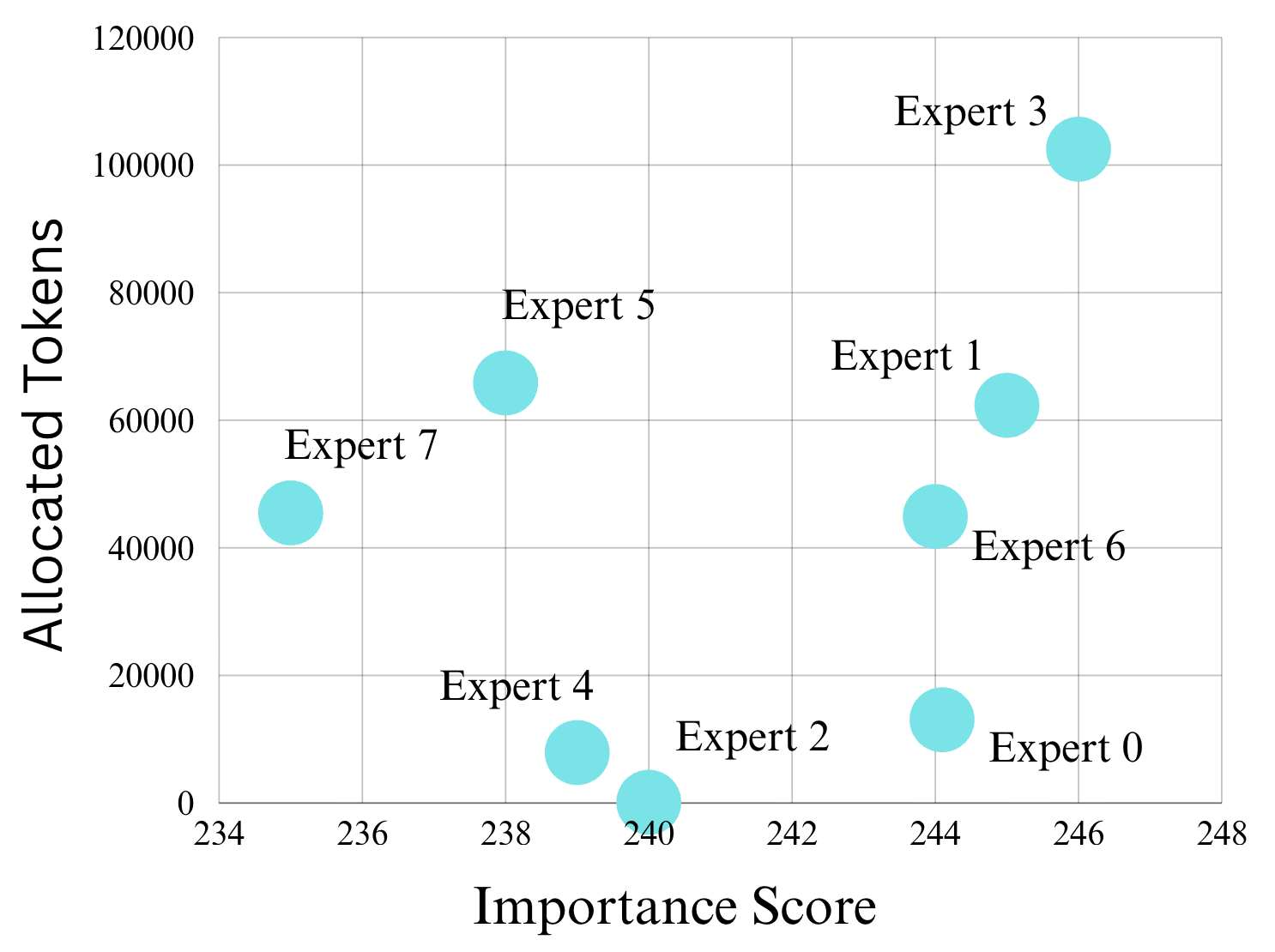}
    \caption{\textbf{Allocated tokens versus the inverse of the pruning-based importance score for experts in the first MoE block of LLaMA-MoE on the PIQA dataset.} Each point represents one expert. The weak correlation between token allocation and inverse importance illustrates that heavily utilized experts are not necessarily the most important for task performance, motivating the decoupling of routing frequency and expert importance in our R\&Q design.}
    \label{fig:hh_important}
    \vspace{-5mm}
\end{wrapfigure}

\textbf{The Gap: Lack of Dynamic, Inference-Time Load Balancing.} 
Current approaches can improve load balance within the training distribution, but they often fail to generalize at inference time—particularly under non-stationary or streaming inputs. Once deployed, the routing policy is typically fixed, leaving little room for dynamic adaptation to changing workloads or system constraints. In practice, inference-time token distributions often diverge significantly from those observed during training. For example, a model trained primarily on general-domain text may later be used for specialized tasks such as biomedical or legal queries, where token statistics and routing dynamics differ substantially. Such distributional shifts are especially common in interactive or continually evolving systems, where user queries change over time. As a result, routing policies tuned for the training distribution can become suboptimal, leading to worsening load imbalance in deployment.

 This challenge has recently attracted attention from the systems community. Capacity-Aware Inference~\cite{he2025capacityawareinferencemitigatingstraggler} and HarMoEny~\cite{doucet2025harmoenyefficientmultigpuinference} identify overused experts as stragglers that degrade latency and GPU utilization during distributed inference. MoEShard~\cite{Dhasade_2025} and GRACE-MoE~\cite{han2025gracemoegroupingreplicationlocalityaware} further propose expert replication and sharding mechanisms to mitigate cross-device imbalance. While effective, these methods typically rely on customized runtimes or architectural changes, limiting their compatibility with existing SMoE models.

 \begin{wrapfigure}{r}{0.5\textwidth}
    \centering
    \vspace{-4mm}
    \includegraphics[width=0.48\textwidth]{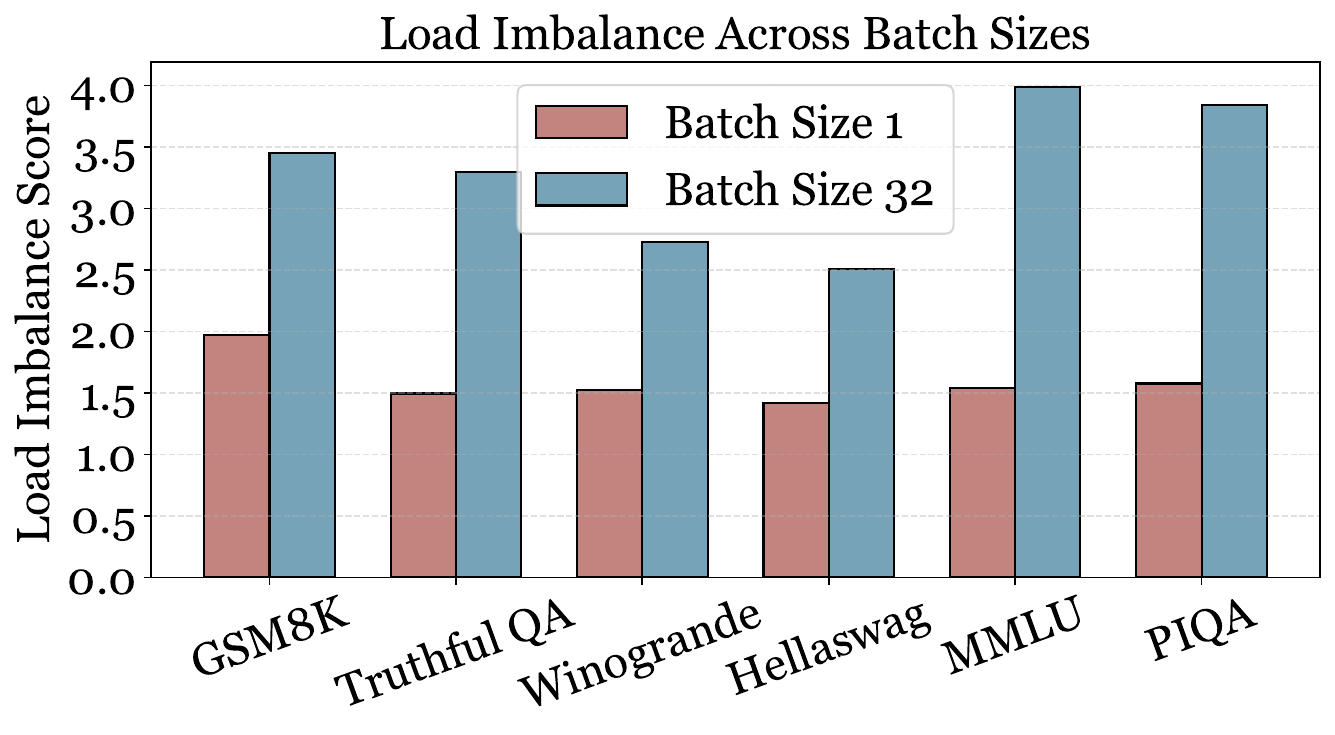}
    \caption{\textbf{Effect of Batch Size on Load Imbalance.} 
    This figure reports the LIS (Definition~\ref{def:lb_score}) for six representative tasks when evaluated under batch sizes of 1 and 32 using the Switch Transformer (8 experts). 
    Across all datasets, larger batch sizes consistently amplify load imbalance, as independently routed tokens increasingly concentrate on a small subset of experts. 
    This highlights the scalability limitation of static routing and underscores the need for inference-time adaptation such as R\&Q to maintain balanced expert utilization.}
    \label{fig:batchsize_performance}
    \vspace{-6mm}
\end{wrapfigure}

Compounding this issue, most SMoE routers are implemented using lightweight functions—such as single-layer MLPs or top-$k$ scoring—that prioritize speed over adaptability. These routers are computationally efficient but static: once trained, they cannot adjust to new input distributions or system-level constraints like memory budgets or latency targets. Consequently, deployed models often experience persistent overload on a small subset of experts, reducing throughput and efficiency. Ultimately, this reveals a key gap between training-time balancing and inference-time deployment: while the former improves fairness during learning, it provides no guarantees under dynamic, real-world workloads. What is missing is an inference-time mechanism that can flexibly rebalance expert workloads without retraining or architectural modification.

\textbf{Our Insight: Expert Frequency not equal to Expert Importance.}  
In this work, we first conduct a detailed empirical study to examine how expert routing frequency correlates with its contribution to model performance. Across multiple SMoE architectures and datasets, we observe a consistent divergence: a small set of experts are activated extremely often, yet many of them contribute little to overall accuracy once their outputs are selectively ablated. Conversely, several infrequently used experts exhibit high marginal utility, suggesting that the gating mechanism tends to favor popularity over usefulness. This mismatch stems partly from static routing functions that reinforce early activation biases and partly from non-stationary input distributions during inference. Together, these factors cause certain experts to become overutilized while others remain underexploited, amplifying inference-time inefficiency. 

This finding carries a key implication for inference optimization. Since heavy-hitter experts dominate latency but not necessarily accuracy, and low-importance experts occupy resources without meaningful contribution, we can decouple two complementary objectives: alleviating overloaded experts and preserving the contribution of important experts. Specifically, experts that are overloaded should be replicated to alleviate computational bottlenecks, while the least important experts can be aggressively quantized to reduce memory and compute usage with minimal accuracy loss. 

\begin{figure*}[t]
    \centering
    \includegraphics[width=1\textwidth]{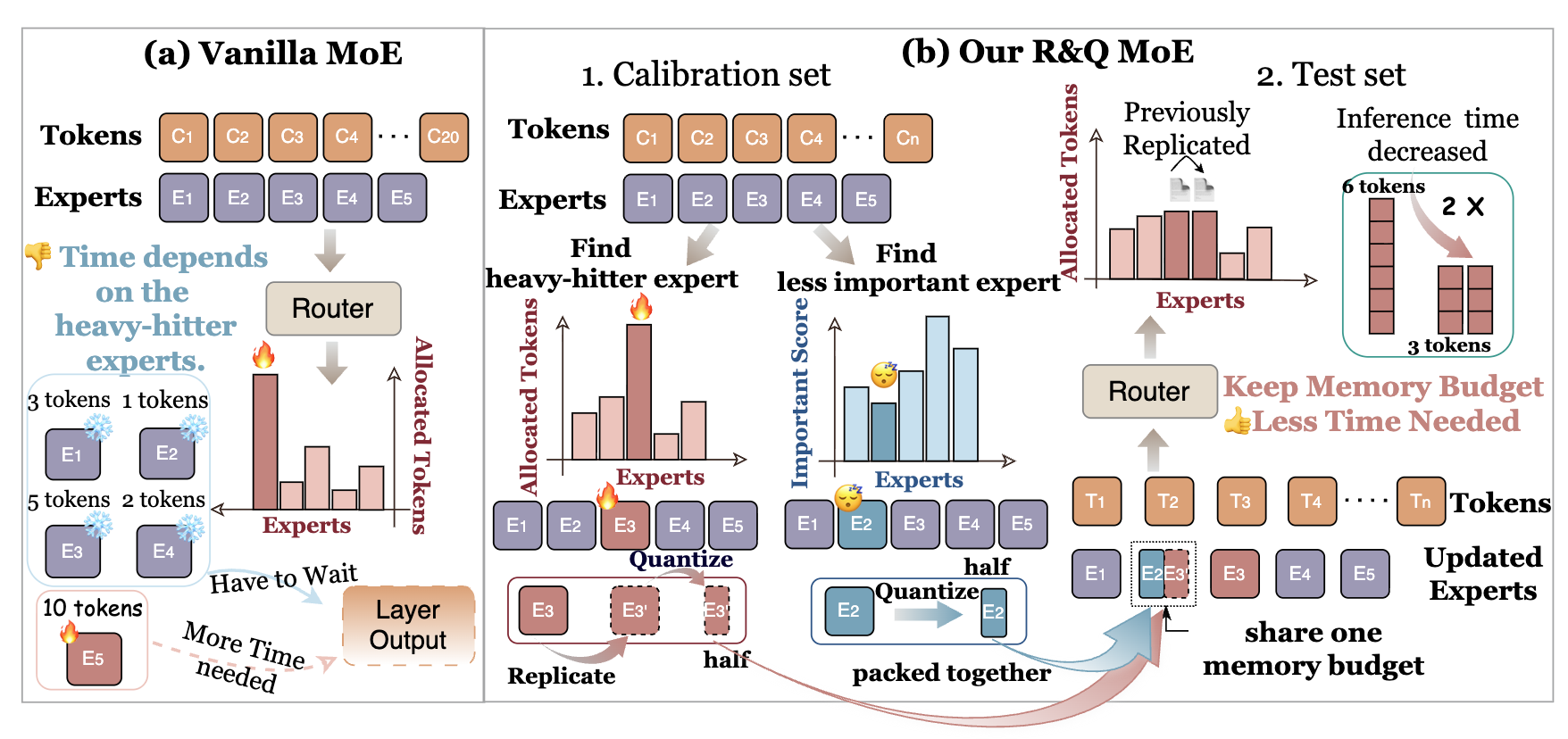}
    \caption{
\textbf{Overview of the Replicate-and-Quantize (R\&Q) framework compared to vanilla MoE.}
(a) Vanilla MoE exhibits inference-time load imbalance, where a small number of heavy-hitter experts receive a disproportionate share of tokens due to the router’s allocation pattern.  
(b) The proposed R\&Q method identifies heavy-hitter and less important experts using a small calibration set, replicates the former, and quantizes both the latter and the replicas under the same memory budget. During inference, the replicated experts mitigate routing bottlenecks and reduce token concentration, leading to more balanced expert utilization without requiring any additional training.
}
\label{fig:pipeline}
\end{figure*}

\textbf{Our Proposal: Replicate-and-Quantize (R\&Q).}  
These complementary analysis motivate our proposed \textbf{Replicate-and-Quantize}, a training-free, plug-and-play framework for mitigating load imbalance in SMoEs during inference. Our method identifies the heaviest-loaded experts and replicates them using lower-bit quantized versions, effectively distributing their computational burden without altering model semantics. Replication provides parallel execution capacity, ensuring that heavy-hitter experts no longer become stragglers in distributed inference pipelines. Quantization, in turn, ensures that the memory and compute budget remains bounded, even as we add replicas. Simultaneously, we quantize underutilized experts to preserve the memory and compute budget. This dual strategy is critical: replication alone would inflate resource requirements, while blanket quantization would degrade accuracy. By combining the two, R\&Q balances efficiency and fidelity. Unlike retraining-based approaches, R\&Q operates post hoc and can be applied directly to pre-trained models, making it suitable for practical deployment scenarios.

To evaluate the effectiveness of our framework, we also introduce a novel metric—\textbf{Load Imbalance Score (LIS)}—that quantitatively measures the skew in expert utilization. This metric allows us to rigorously evaluate the impact of routing strategies and validate the effectiveness of our proposed framework.

\textbf{Contributions.}  
In summary, this work makes the following key contributions:

\begin{itemize}[leftmargin=*]
    \item We conduct a comprehensive empirical analysis of expert selection patterns in SMoEs, revealing that selection frequency does not reliably indicate an expert’s true utility. This insight highlights a misalignment between routing dynamics and expert utility, motivating a design that explicitly decouples expert load balance from utility preservation at inference time.
    \item We introduce the \textbf{Load Imbalance Score (LIS)}, a metric that quantitatively reflects the imbalance of expert utilization in SMoEs by measuring how far the current routing deviates from ideal balanced allocation. LIS enables rigorous and comparable evaluation of routing strategies, providing a principled basis for assessing inference-time load balancing approaches. 
    \item Based on LIS, we propose \textbf{Replicate-and-Quantize}, a lightweight, training-free method for SMoE inference that dynamically balances expert workloads through replication and quantization, achieving more balanced and efficient inference without requiring retraining.
    \item We validate our approach across five model architectures and six datasets in both static and streaming inference scenarios. Experiments demonstrate that R\&Q significantly reduces load imbalance by up to \textbf{1.4x}, while maintaining model performance within a margin of \textbf{±0.6\%} accuracy across all tasks, highlighting its robustness and practicality for diverse deployment environments.
\end{itemize}

Together, these contributions close a crucial gap in the SMoE literature by shifting the focus from training-time routing optimization to practical, inference-time adaptability. We believe our findings pave the way for more robust and deployable SMoE-based systems in diverse application domains.

\begin{table*}[t]
\centering
\small
\setlength{\tabcolsep}{5pt}
\renewcommand{\arraystretch}{1.12}
\resizebox{\textwidth}{!}{%
\begin{tabular}{lcccccccc|c}
\toprule
\textbf{Method} &
GSM8K & MMLU & TruthfulQA & PIQA & WikiQA & Hellaswag & Winogrande & \textbf{Avg LIS} & \textbf{WikiQA Acc. (\%)} \\
\midrule
Tune router 2nd   & 2.4556 & 2.8854 & 2.4165 & 2.4814 & 2.7746 & 2.6985 & 2.8346 & 2.6495 & 20.39 \\
Freeze router     & 1.8976 & 1.6155 & 1.4866 & 1.5417 & 1.3630 & 1.3925 & 1.4697 & 1.5381 & 20.35 \\
Tune router 10ep  & 2.4935 & 2.7618 & 2.3970 & 2.1930 & 2.8080 & 2.4694 & 2.8425 & 2.5665 & 21.33 \\
Tune both         & 2.3379 & 3.2713 & 2.8660 & 2.6816 & 3.2152 & 2.8884 & 3.1544 & 2.9164 & 20.62 \\
Freeze router 1ep & 2.1052 & 1.8082 & 1.6714 & 1.5206 & 1.5710 & 1.5718 & 1.7276 & 1.7108 & 20.57 \\
Tune router       & 2.7509 & 3.7665 & 3.5322 & 3.3336 & 3.9062 & 3.3334 & 4.1324 & 3.6793 & 11.35 \\
Tune expert       & 1.9614 & 1.7298 & 1.6715 & 1.6909 & 1.5011 & 1.6527 & 1.6455 & 1.6933 & 19.55 \\
Full finetune     & 2.3056 & 2.7234 & 2.4218 & 2.1323 & 2.6271 & 2.3064 & 2.6285 & 2.4493 & 20.11 \\
Original          & 1.9709 & 1.5405 & 1.4956 & 1.5770 & 1.3910 & 1.4182 & 1.5261 & 1.5599 & 13.96 \\
\rowcolor[HTML]{EAEAEA}
\textbf{Replicate-and-Quantize (R\&Q)} &
\textbf{1.3937} & \textbf{1.2962} & \textbf{1.3494} & \textbf{1.2756} & \textbf{1.2864} & \textbf{1.3623} & \textbf{1.2146} & \textbf{1.3112} & \textbf{19.35} \\
\bottomrule
\end{tabular}%
}
\caption{
\textbf{Effect of fine-tuning strategies on load balancing and accuracy.}
We report the LIS (Definition~\ref{def:lb_score}) across seven tasks, their average (Avg LIS), and WikiQA accuracy. 
Router-aggressive fine-tuning slightly improves accuracy but causes severe imbalance, whereas router-frozen variants reduce LIS with limited accuracy gain. 
Our Replicate-and-Quantize (R\&Q) method achieves the lowest Avg LIS while maintaining comparable accuracy, demonstrating superior inference-time load balancing without retraining.
}
\label{tab:hardness_finetune}
\end{table*}

\section{Method: Replicate-and-Quantize for Efficient SMoE Inference}
As illustrated in Figure~\ref{fig:moe_small}, our proposed \textbf{Replicate-and-Quantize (R\&Q)}, a training-free, plug-and-play framework to address inference-time load imbalance in Sparse Mixture-of-Experts (SMoE) models. Unlike training-time regularization approaches~\cite{lepikhin2020gshard, fedus2022switch, zoph2022stmoedesigningstabletransferable}, R\&Q operates purely at inference and remains robust under diverse deployment settings such as streaming inputs. This section formalizes the load imbalance phenomenon, introduces a lightweight metric to quantify it, and presents our post-hoc adaptation algorithm grounded in observed expert utilization patterns.

We begin by introducing the key notations and definitions relevant to SMoE models. A Sparse Mixture-of-Experts (SMoE) model consists of multiple MoE blocks, each containing $m$ parallel expert networks. For each input token, a lightweight routing function selects the top-$k$ ($k < m$) experts based on token-specific scores. This selective activation enables computational savings but often leads to uneven expert utilization.

To formalize this, let $n_{i,j}$ denote the number of tokens routed to expert $j$ in MoE block $i$. The total number of expert selections for block $i$ is $nk$, where $n$ is the total number of input tokens. To quantitatively characterize such uneven expert utilization, we introduce the following lightweight metric that measures the degree of expert load skewness:

\begin{definition}[Load Imbalance Score]\label{def:lb_score}
Let $\mathcal{M}$ be a SMoE model with $p$ MoE blocks and $m$ experts per block. Each token activates $k < m$ experts. For block $i$, let $n_{i,j}$ be the number of tokens routed to expert $j$. The LIS for block $i$ is defined as:
\begin{align*}
    l_i = \frac{m \cdot \max_{j\in [m]} n_{i,j}}{nk}.
\end{align*}
This score reflects how much the busiest expert deviates from ideal balanced routing. A value of 1 indicates perfect balance, while larger values imply growing imbalance.
\end{definition}

\subsection{Motivation and Challenges}
Based on LIS, we empirically analyze expert utilization patterns in pretrained SMoE models and find that inference-time load imbalance persists even when the router was trained with load-balancing objectives. 
In particular, as batch size increases or input distributions shift, a few heavy-hitter experts dominate the routing, leading to latency bottlenecks and inefficient hardware utilization. 
While SMoEs achieve computational efficiency by activating only a subset of experts per token, this imbalance severely limits their scalability at inference. 
Prior work has proposed training-time regularization like auxiliary balancing losses to address this issue~\cite{lepikhin2020gshard, fedus2022switch, zoph2022stmoedesigningstabletransferable}, but these methods often fail to generalize to unseen or dynamically changing inputs during deployment. 
This motivates an inference-time adaptation mechanism that can rebalance workloads without retraining or altering the router.

\begin{wrapfigure}{l}{0.52\textwidth}
\vspace{-1em}
\begin{minipage}{0.50\textwidth}
\captionsetup{hypcap=false}
\begin{algorithm}[H]
\caption{Search for Heavy-Hitter Expert}
\label{alg:heavy_expert}
\small
\begin{algorithmic}
\State \textbf{Input:}
\State \hspace{1em} $X$ = Input tokens
\State \hspace{1em} $En$ = Expert numbers
\State \hspace{1em} $L$ = MoE Layers
\State \hspace{1em} $T$ = Token numbers
\State \hspace{1em} $s$ = Sparsity factor
\State \textbf{Output:} List of Heavy Experts $EC$
\State \textbf{Initialize:} $EC \leftarrow \text{list}[L]$
\For{$l \in L$}
    \State $\texttt{expert\_chosen} \leftarrow [\,]$
    \For{$x \in X$}
        \State $\texttt{expert\_chosen} \leftarrow l(x)$
    \EndFor
    \State $\texttt{expert\_num} \leftarrow \text{count}(\texttt{expert\_chosen})$
    \State $\texttt{heavy\_expert} \leftarrow \arg\max(\texttt{expert\_num})$
    \State $EC[l] \leftarrow \texttt{heavy\_expert}$
\EndFor
\Return $EC$
\end{algorithmic}
\end{algorithm}
\end{minipage}

\end{wrapfigure}

\noindent\textbf{Challenge 1: Inference-Time Imbalance is Not Solved by Fine-Tuning.}
While many prior efforts focus on training-time strategies to promote expert utilization, we empirically demonstrate that post-hoc fine-tuning struggles to improve load balancing without harming task performance. To illustrate this, we adopt the Switch Transformer~\cite{fedus2022switch} and conduct a series of fine-tuning experiments using the load-balancing loss proposed in the original work. Specifically, we evaluate multiple strategies, including \textbf{Full Finetune}, where all model parameters are updated; \textbf{Tune Expert}, where only expert weights are updated; and \textbf{Tune Router}, which updates only router weights. We also consider hybrid or staged strategies: \textbf{Freeze Router 1ep}, which freezes router weights during the first epoch before releasing them for full finetuning; \textbf{Tune Both}, which jointly tunes router and expert weights; and \textbf{Tune Router 10ep}, where the router is tuned for 10\% of training steps in each epoch before tuning all weights. In addition, we evaluate \textbf{Freeze Router}, which keeps router weights frozen while tuning all other parameters, and \textbf{Tune Router 2nd}, which first tunes the full model and then tunes only the router during the final two epochs.

Table~\ref{tab:hardness_finetune} reports LIS (as defined in Definition~\ref{def:lb_score}) on seven tasks and accuracy on WikiQA. Fine-tuning strategies that aggressively update the router, such as Tune Router 10ep, Tune Both, or Full Finetune, achieve the highest accuracy on WikiQA (around 21\%) but also produce substantially higher imbalance scores—for example, the LIS on MMLU increases to over 3.2. In contrast, methods that restrict router updates, including Freeze Router and Tune Expert, reduce the imbalance scores but bring little or no improvement in accuracy. This pattern reveals a clear trade-off between accuracy and expert load balance: improving one often deteriorates the other.

\noindent\textbf{Challenge 2: Imbalance Worsens with Increased Batch Size.}
In addition to the challenges of post-hoc fine-tuning, we observe that load imbalance in Sparse Mixture-of-Experts models worsens consistently as inference batch size increases. As shown in Figure~\ref{fig:batchsize_performance}, the LIS increases significantly—for example, rising from 1.5405 to 3.9897 on MMLU as batch size grows from 1 to 32. This trend is consistent across a variety of tasks and model settings.

The underlying cause of this pattern lies in the design of typical routing mechanisms. Since most routing functions operate independently for each token and do not coordinate assignments globally, imbalanced expert usage in large batches tends to accumulate. This phenomenon has practical implications: modern deployments often prefer larger batch sizes to improve hardware utilization and throughput. Without addressing load imbalance at inference time, such deployments may suffer from degraded efficiency, expert overload, or device under-utilization. These observations motivate the need for a solution that adapts expert assignment dynamically during inference, without retraining or structural changes to the model.

\begin{wrapfigure}{r}{0.52\textwidth}
\vspace{-1em}
\begin{minipage}{0.50\textwidth}
\captionsetup{hypcap=false}
\begin{algorithm}[H]
\caption{Search for Less Important Expert}
\label{alg:less_important_expert}
\begin{algorithmic}
\State \textbf{Input:} Layer inputs $X$, experts' weights $W$, MoE layers $L$, number of experts $E$
\State \textbf{Output:} List of Less Important Experts $IE$
\State \textbf{Initialize:}
\State \hspace{1em} $IE \leftarrow \text{list}[L]$
\State \hspace{1em} $S \leftarrow \text{list}[L][E]$
\State \hspace{1em} $IS \leftarrow \text{list}[L][E]$
\State Extract less important experts $e_1, e_2, \dots, e_n$ for each MoE layer $L$ based on inputs $X$
\For{$l \in L$}
    \For{$e \in E$}
        \State $S_{[l][e]} \leftarrow \left| W_{[l][e]}^{T} \right| \times \|X_{[l]}\|_{2}$
        \State $\texttt{sorted\_idx}_{[l][e]} \leftarrow 
        \texttt{sort}(S_{[l][e]})$ \Comment{descending}
        \State $\texttt{pruned\_idx}_{[l][e]} \leftarrow
        \texttt{sorted\_idx}_{[l][1:\text{int}(C_{\text{in}} \times s)]}$
        \State $\texttt{expert\_score} \leftarrow
        \texttt{mean}\big(S_{\texttt{pruned\_idx}_{[l][e]}}\big)$
        \State $IS_{[l][e]} \leftarrow
        IS_{[l][e]} + \texttt{expert\_score}$
    \EndFor
    \State $IE_{l} \leftarrow
    \arg\max_{e}\ IS_{[l][e]}$
\EndFor
\Return $IE$
\end{algorithmic}
\end{algorithm}
\end{minipage}
\vspace{-1em}
\end{wrapfigure}

\noindent\textbf{Summary.}
The above two challenges highlight that inference-time load imbalance in Sparse Mixture-of-Experts models cannot be effectively addressed by existing training-time or fine-tuning approaches. Fine-tuning methods tend to trade accuracy for balance, and vice versa, while increasing the inference batch size further amplifies the imbalance due to independent token-level routing. Together, these findings motivate the need for an inference-time mechanism that can dynamically balance expert utilization across varying batch conditions, without relying on retraining or altering the original model design. To design such a mechanism, it is essential to understand not only which experts are overloaded but also which ones are truly important for model performance. In the following, we first present a case study that examines the relationship between expert load and importance, revealing why both dimensions must be considered jointly in addressing inference-time imbalance.

\subsection{Heavy-Hitter vs. Importance: A Case Study}
To motivate the need for a more nuanced inference-time adaptation mechanism, we analyze the relationship between expert load and expert importance in Sparse Mixture-of-Experts models. Specifically, we collect routing statistics and per-expert importance scores from a pretrained LLaMA-MoE model evaluated on the PIQA dataset. For each expert within the first MoE block, we record the total number of tokens routed to it as an estimate of its load. The corresponding importance scores are obtained using the Wanda~\cite{sun2024simpleeffectivepruningapproach} metric, which estimates an expert’s contribution to model accuracy in a gradient-free manner. Figure~\ref{fig:hh_important} visualizes the relationship between how frequently each expert is selected during routing and how critical it is to model performance.

Empirically, we observe that expert load and importance exhibit weak or no clear correlation: some heavily utilized experts contribute little to overall task accuracy, whereas certain rarely activated ones are critical. This suggests that load balancing and expert significance are largely decoupled aspects of MoE behavior. Consequently, addressing inference-time imbalance requires more than merely redistributing token traffic—it must also account for which experts truly matter for prediction quality.

Building on this insight, we decompose the inference-time adaptation problem into two complementary dimensions: (i) identifying and mitigating routing bottlenecks caused by heavy-hitter experts (Section~\ref{sec:hh}), and (ii) quantifying expert importance to guide selective compression (Section~\ref{sec:importance}). Together, these analyses form the foundation for our proposed Replicate-and-Quantize (R\&Q) framework (Section~\ref{sec:rq}).

\subsection{Identifying Heavy-Hitter Experts in SMoE} \label{sec:hh}
Due to the inherent design of routing mechanisms in Sparse Mixture-of-Experts (SMoE) models, a small subset of experts naturally receive a disproportionate amount of token traffic. In most SMoE architectures, each token is routed independently to its top-$k$ experts based on local gating probabilities without any coordination across tokens. This independence means that if certain experts consistently obtain slightly higher routing scores, even marginal differences are amplified across large batches of tokens. In practice, the gating softmax tends to produce peaked distributions, so a few experts dominate the top-$k$ selections. Moreover, token distributions in natural language or multimodal inputs are highly non-uniform—tokens with similar semantics or frequency often activate the same experts. Together, these factors cause some experts to become “heavy-hitters,” processing far more tokens than others. Such imbalance not only creates computational bottlenecks but also leads to inefficient expert utilization.

We define this phenomenon formally as follows:
\begin{oracle}[Heavy-Hitter Expert]\label{oracle:hh}
    Let $\mathcal{M}$ denote a sparse mixture-of-experts (SMoE) model with $p$ MoE blocks. Each block contains $m$ expert networks, and each token activates $k < m$ experts. Given an input dataset $X$ of $n$ tokens, we count the number of tokens assigned to expert $j$ at block $i$ as $n_{i,j}$. The expert $j^*$ with the maximum $n_{i,j}$ in block $i$ is designated the heavy-hitter expert:
    $$j^* = \arg\max_{j \in [m]} n_{i,j}.$$
\end{oracle}

\begin{wrapfigure}{r}{0.5\textwidth}
\vspace{-8pt}
\begin{minipage}{0.48\textwidth}
\centering
\small
\setlength{\tabcolsep}{5pt}
\renewcommand{\arraystretch}{1.12}

\resizebox{\textwidth}{!}{%
\begin{tabular}{lccccc}
\toprule
\textbf{Model} & \textbf{Task} & \textbf{Ours} & Random & Heavy-hitter & Raw \\ 
\midrule
\multirow{6}{*}{\textbf{LLaMA-MoE}} 
 & GSM8K       & \textbf{4.17\%} & 3.64\% & 2.96\% & 4.25\% \\
 & Hellaswag   & \textbf{51.79\%} & 51.66\% & 51.26\% & 54.14\% \\
 & MMLU        & 26.69\% & 26.39\% & \textbf{26.81\%} & 27.81\% \\
 & PIQA        & \textbf{75.79\%} & 75.35\% & 75.24\% & 76.93\% \\
 & TruthfulQA  & 28.64\% & \textbf{30.01\%} & 28.64\% & 27.26\% \\
 & Winogrande  & \textbf{64.80\%} & 55.72\% & 63.69\% & 67.01\% \\ 
\midrule
\multirow{6}{*}{\textbf{\shortstack{Switch Transformer\\(8 Experts)}}} 
 & GSM8K       & 0.45\% & \textbf{1.06\%} & 0.38\% & 0.00\% \\
 & Hellaswag   & \textbf{28.26\%} & 28.21\% & 28.25\% & 27.46\% \\
 & MMLU        & \textbf{22.95\%} & 22.95\% & 22.95\% & 22.95\% \\
 & PIQA        & \textbf{59.41\%} & 58.43\% & 59.25\% & 58.11\% \\
 & TruthfulQA  & 37.56\% & 36.14\% & \textbf{37.92\%} & 36.92\% \\
 & Winogrande  & \textbf{54.54\%} & 51.93\% & 52.25\% & 49.64\% \\ 
\bottomrule
\end{tabular}
}

\captionsetup{type=table}
\caption{
\textbf{Evaluating expert removal strategies in SMoEs.}
Accuracy (\%) after removing one expert per layer, comparing our Wanda-based selection (“Ours”) against Random and Heavy-hitter baselines, with Raw as the unmodified model.  
Across both LLaMA-MoE and Switch Transformer (8 Experts), our method consistently preserves or improves accuracy, indicating more reliable identification of less-important experts.  
This confirms that importance-aware pruning maintains task fidelity while offering additional compression headroom.
}
\label{tab:less_important}

\end{minipage}
\vspace{-10mm}
\end{wrapfigure}

We track token-to-expert assignments using inference-time routing traces and compute the token count per expert in each block. Empirically, we find that using only 10\% of the input data is sufficient to robustly estimate heavy-hitter experts, offering a low-overhead diagnostic. We further validate this observation on the Switch Transformer with 16 experts, where the estimated set of heavy-hitters covers over 99\% of all expert selections. The exact steps are detailed in Algorithm~\ref{alg:heavy_expert}.

\subsection{Quantifying Expert Importance in SMoE}
\label{sec:importance}

While heavy-hitter experts are often overloaded, they are not necessarily the most critical to model performance. To assess the importance of each expert, we adopt a gradient-free, pruning-based metric inspired by Wanda that does not rely on backpropagation or fine-tuning.

Let $W \in \mathbb{R}^{C_{\text{out}} \times C_{\text{in}}}$ be the expert's weight matrix and $X \in \mathbb{R}^{(N \cdot L) \times C_{\text{in}}}$ denote the corresponding input activations. We compute the $L_2$ norm of $X$ across the batch and sequence dimensions:
$$\|X\|_2 = \sqrt{\sum_{i=1}^{N \cdot L} X_{ij}^2} \quad \text{for each column } j.$$ 

Next, we compute an element-wise Wanda score $S$ by multiplying $|W|$ with $\|X\|_2$. We sort the scores in ascending order and prune the bottom-$s$ fraction, corresponding to the least impactful weights. For each expert, we average the pruned scores to obtain an importance proxy—lower means imply higher importance.

This procedure is outlined in Algorithm~\ref{alg:less_important_expert}. To validate this metric, we compare against two baselines: (1) randomly selected one expert as less important one and (2) heavy-hitter experts. As shown in Table~\ref{tab:less_important}, our Wanda-based metric more reliably identifies experts whose removal causes minimal degradation in task accuracy. 

\subsection{Our Proposal: Replicate-and-Quantize} \label{sec:rq}
We propose a training-free, plug-and-play strategy called \textbf{Replicate-and-Quantize (R\&Q)} to mitigate inference-time load imbalance in Sparse Mixture-of-Experts (SMoE) models. Unlike approaches that modify the training process or router architecture, R\&Q operates entirely at inference time. It does so by restructuring the expert topology through selective replication and quantization of experts, as illustrated in Figure~\ref{fig:pipeline}.
R\&Q comprises two main components:

\begin{wrapfigure}{r}{0.52\textwidth}
\vspace{-1em}
\begin{minipage}{0.50\textwidth}
\captionsetup{hypcap=false}
\begin{algorithm}[H]
\caption{Replicate-and-Quantize Algorithm}
\label{alg:qd}
\begin{algorithmic}
\State \textbf{Input:} Model $M$, number of experts $E$
\State \textbf{Output:} Replicated and quantized model $RQ$
\State \textbf{Initialize:}
\State \hspace{1em} $\texttt{replicate\_expert}
\leftarrow$ Algorithm~\ref{alg:heavy_expert}
\State \hspace{1em} $\texttt{quantization\_expert}
\leftarrow$ Algorithm~\ref{alg:less_important_expert}
\State $\texttt{layer\_idx} \leftarrow 0$
\For{$\texttt{layer} \in M$}
    \State $\texttt{re} \leftarrow
    \texttt{replicate\_expert}_{[\texttt{layer\_idx}]}$
    \State $\texttt{qe} \leftarrow
    \texttt{quantization\_expert}_{[\texttt{layer\_idx}]}$
    \State $\texttt{layer}
    \leftarrow
    \texttt{layer}
    + \texttt{quant}(\texttt{layer}_{[\texttt{re}]})$
    \State $\texttt{layer}_{[\texttt{qe}]}
    \leftarrow
    \texttt{quant}(\texttt{layer}_{[\texttt{qe}]})$
    \State $\texttt{layer\_idx}
    \leftarrow
    \texttt{layer\_idx} + 1$
\EndFor
\Return $M$
\end{algorithmic}
\end{algorithm}
\end{minipage}
\vspace{-1mm}
\end{wrapfigure}

\noindent\textbf{Replicating Heavy-Hitter Experts.}
We begin by identifying experts that are consistently overutilized—referred to as heavy-hitters—based on token routing patterns collected from a representative inference dataset. As described in Algorithm~\ref{alg:heavy_expert}, we select the most frequently invoked expert within each MoE block. To alleviate the computational bottleneck caused by such experts, we replicate them within the same block. These replicated experts retain the same parameters but are deployed at lower precision (e.g., 8-bit quantization) to reduce their memory footprint. Importantly, this replication does not require any modification to the model’s routing mechanism.

\noindent\textbf{Quantizing Less Important Experts.}
In parallel, we identify experts that contribute minimally to overall task performance. These less important experts are determined using an importance score described in Section~\ref{sec:importance}, which draws inspiration from pruning methods such as Wanda. As shown in Algorithm~\ref{alg:less_important_expert}, experts that exhibit high parameter redundancy—i.e., those whose weights can be significantly pruned with little impact on performance—are considered suitable candidates. These experts are then quantized more aggressively, freeing up memory and compute resources. This allows the added replicas of heavy-hitter experts to fit within the model’s existing resource constraints, all without compromising accuracy.

Together, these two components form a unified strategy for inference-time adaptation that is applicable to a wide range of SMoE architectures without retraining or architecture redesign.
The overall Replicate-and-Quantize (R\&Q) workflow is summarized in Algorithm~\ref{alg:qd}.

\section{Experiments}
In this section, We evaluate the effectiveness of R\&Q across models and tasks, answering the following research questions. 

\subsection{Preliminary Setups}
\textbf{Dataset.} In this work, we evaluate the performance of our proposed method using a diverse set of benchmark datasets to test different aspects of model performance across a variety of domains and task types. The datasets include Massive Multitask Language Understanding (MMLU)~\cite{hendrycks2021measuringmassivemultitasklanguage}, TruthfulQA~\cite{lin2022truthfulqameasuringmodelsmimic}, Grade School Math (GSM8K)~\cite{cobbe2021trainingverifierssolvemath}, Winogrande~\cite{sakaguchi2019winograndeadversarialwinogradschema}, Hellaswag~\cite{zellers2019hellaswagmachinereallyfinish}, and Physical Interaction Question Answering (PIQA)~\cite{bisk2019piqareasoningphysicalcommonsense}. For the fine-tuning experiment, we tuned the model using the WikiQA~\cite{yang-etal-2015-wikiqa} dataset, a public question-answering benchmark focused on the quality of Wikipedia content. During the evaluation of the fine-tuned model, we incorporate the WikiQA dataset into the existing evaluation datasets.

\textbf{Testbed.} We fine-tuned the Switch Transformer model on one NVIDIA Tesla V100-SXM2-32GB, and all Switch Transformers evaluation experiments were conducted on the NVIDIA 8 Tesla V100-SXM2-32GB U servers. The LLaMa-MoE, DeepSeek-MoE and DeepSeek-V2 Lite were evaluated on the NVIDIA 4 A100-SXM4-40GB and NVIDIA 8 Quadro RTX 8000. 

\textbf{Models.}
We adopt three model architectures in our approach: Switch Transformer, LLaMA-MoE, and DeepSeek-MoE. Below is a breakdown of these architectures:

\begin{itemize}
    \item \textbf{Switch Transformer (8 Experts) and Switch Transformer (16 Experts):}  
    Each layer contains 8 experts and 16 experts separately, with only 1 expert selected per token.
    
    \item \textbf{LLaMA-MoE (8 Experts):}  
    Each layer contains 8 experts, with 2 experts selected per token.

    \item \textbf{DeepSeek-MoE and DeepSeek v2 Lite (64 experts):}  
    Each layer consists of 66 experts: 64 isolated experts and 2 shared experts. During routing, the system selects 6 experts from the 64 isolated experts while always utilizing the shared experts. Thus, our strategy is exclusively applied to the 64 isolated experts.
\end{itemize}

\begin{wrapfigure}{r}{0.48\textwidth}
\vspace{-10pt}

\begin{minipage}{0.46\textwidth}
\centering
\small
\setlength{\tabcolsep}{6pt}
\renewcommand{\arraystretch}{1.1}

\captionsetup{type=table}

\begin{tabular}{@{}p{4.2cm}p{2.4cm}@{}}
\toprule
\textbf{Parameter} & \textbf{Value} \\ 
\midrule

Learning rate              & 5e-5  \\
Training epochs            & 10    \\
Training batch size        & 8     \\
Batch size (WikiQA accuracy) & 16 \\
Batch size (LIS computation) & 1  \\
Weight decay               & 0.01  \\
Optimizer                  & AdamW \\

\bottomrule
\end{tabular}
\caption{
\textbf{Fine-tuning hyperparameters for the Switch Transformer (8 Experts).}
These configurations are used for all results in Table~\ref{tab:hardness_finetune}, also including evaluation for both WikiQA accuracy and LIS computation.
}
\label{tab:tuning}

\end{minipage}
\vspace{-10pt}
\end{wrapfigure}

\textbf{Quantization Techniques.}
For Switch Transformer, the model is loaded in \texttt{float32}, so we apply half-precision quantization using \texttt{float16}. For LLaMA-MoE and DeepSeek-MoE, which are loaded in \texttt{float16}, we apply 8-bit quantization as described in \cite{dettmers2022llmint88bitmatrixmultiplication}. This ensures consistent precision reduction across models while adapting to their respective default formats.

\textbf{Implementation Details.}All fine-tuning variants in Table~\ref{tab:hardness_finetune} are evaluated using the Switch Transformer (8 experts) with identical hyperparameters. 
Inference is performed with temperature~1.0, top-k~50, top-p~1.0, and token healing disabled, for both task accuracy and LIS estimation. 
However, for task accuracy on WikiQA, generation is executed with a batch size of~16, whereas the LIS (as defined in Definition~\ref{def:lb_score}) are computed with a batch size of~1.

When computing the LIS, we set \texttt{max\_new\_tokens} to~1.
This setting isolates the routing behavior of the model during a single next-token prediction step, 
allowing us to measure the intrinsic imbalance of expert utilization without confounding effects from multi-step generation dynamics or variable output lengths.

Although only one token is generated per input, evaluating across multiple datasets (e.g., MMLU, PIQA, WikiQA) exposes how different input distributions induce distinct routing patterns, providing a fair and consistent comparison of load imbalance across tasks.
In contrast, WikiQA accuracy is evaluated under standard generation conditions without restricting \texttt{max\_new\_tokens}, reflecting the model’s actual task-level performance.

\textbf{Evaluation Metric.}
We use lm-eval-harness to calculate model accuracy for Massive Multitask Language Understanding (MMLU), TruthfulQA, Grade School Math (GSM8K), Winogrande, Hellaswag, and Physical Interaction Question Answering (PIQA).For the Hellaswag datasets, we use a 10-shot approach, while the GSM8K and MMLU datasets use a 5-shot approach. In our evaluation of the fine-tuned Switch Transformer model, we use the generation F1 score as our criterion.

\begin{wrapfigure}{l}{0.52\textwidth}
\vspace{-2mm}

\centering
\includegraphics[width=0.50\textwidth]{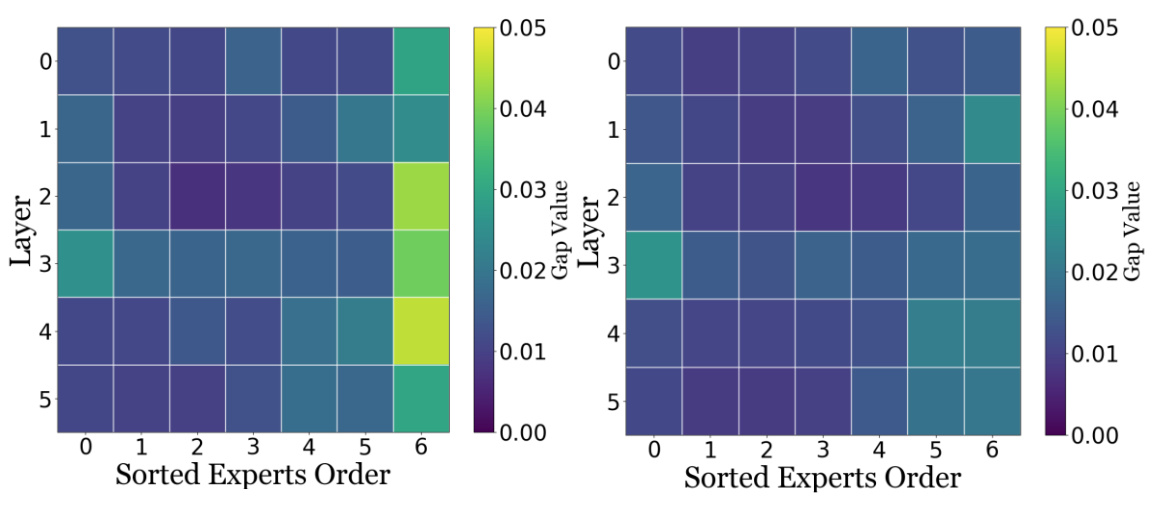}

\caption{\textbf{Per-layer Expert Gap Comparison between R\&Q and the Raw Model.} 
Each heatmap visualizes the normalized gap value (Definition~\ref{def:lb_score}) for all experts across six layers of the Switch Transformer. 
The R\&Q model (left) exhibits a more uniform distribution of gap values across experts, indicating balanced expert utilization, 
whereas the raw model (right) shows higher concentration in specific experts (brighter regions), revealing severe load imbalance.}

\label{fig:combined}
\vspace{-8mm}
\end{wrapfigure}

\textbf{Workload.}
Within each layer, we begin by computing the distribution of tokens across all experts to determine their respective loads. The expert receiving the highest number of tokens is identified as "the most heavy expert", indicating its predominant role in that layer. Concurrently, we utilize the tool "Wanda" ~\cite{sun2024simpleeffectivepruningapproach} to identify the less important expert, characterized by its minimal contribution as assessed by Wanda.

\begin{figure*}[ht!]
    \centering

    \begin{subfigure}{0.24\textwidth}
        \centering
        \includegraphics[width=\linewidth]{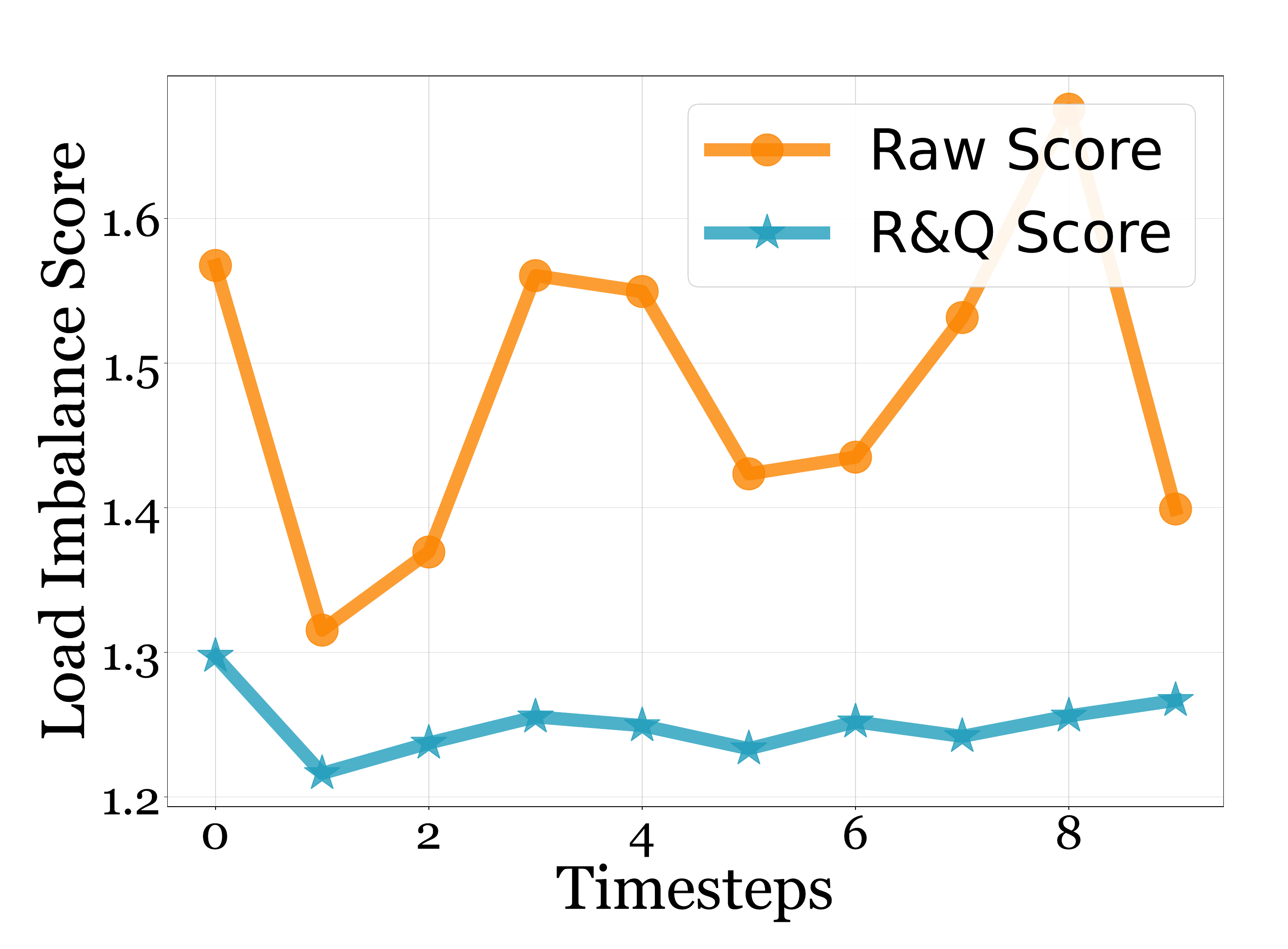}
        \caption{Layer 1}
    \end{subfigure}
    \hfill
    \begin{subfigure}{0.24\textwidth}
        \centering
        \includegraphics[width=\linewidth]{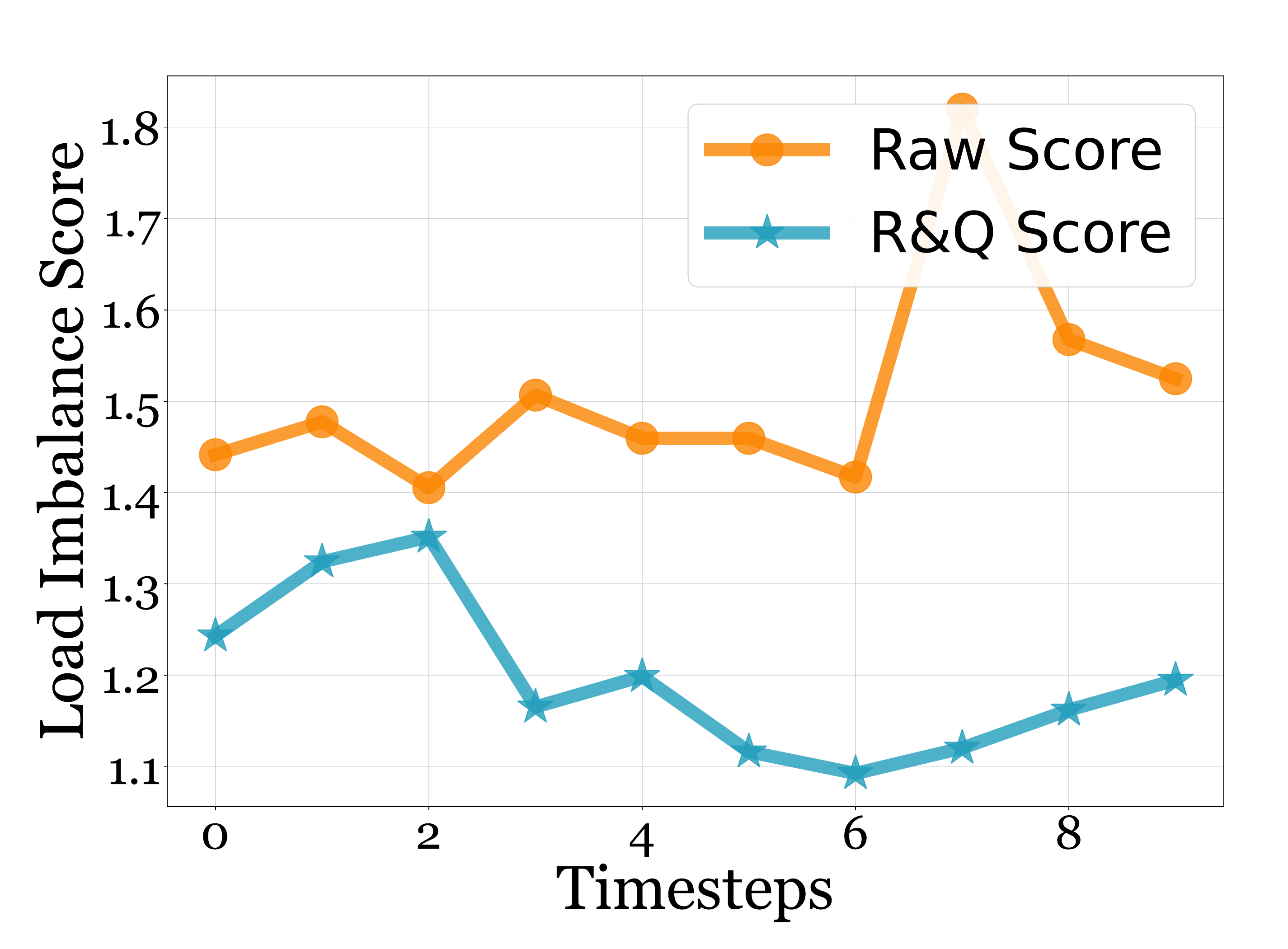}
        \caption{Layer 4}
    \end{subfigure}
    \hfill
    \begin{subfigure}{0.24\textwidth}
        \centering
        \includegraphics[width=\linewidth]{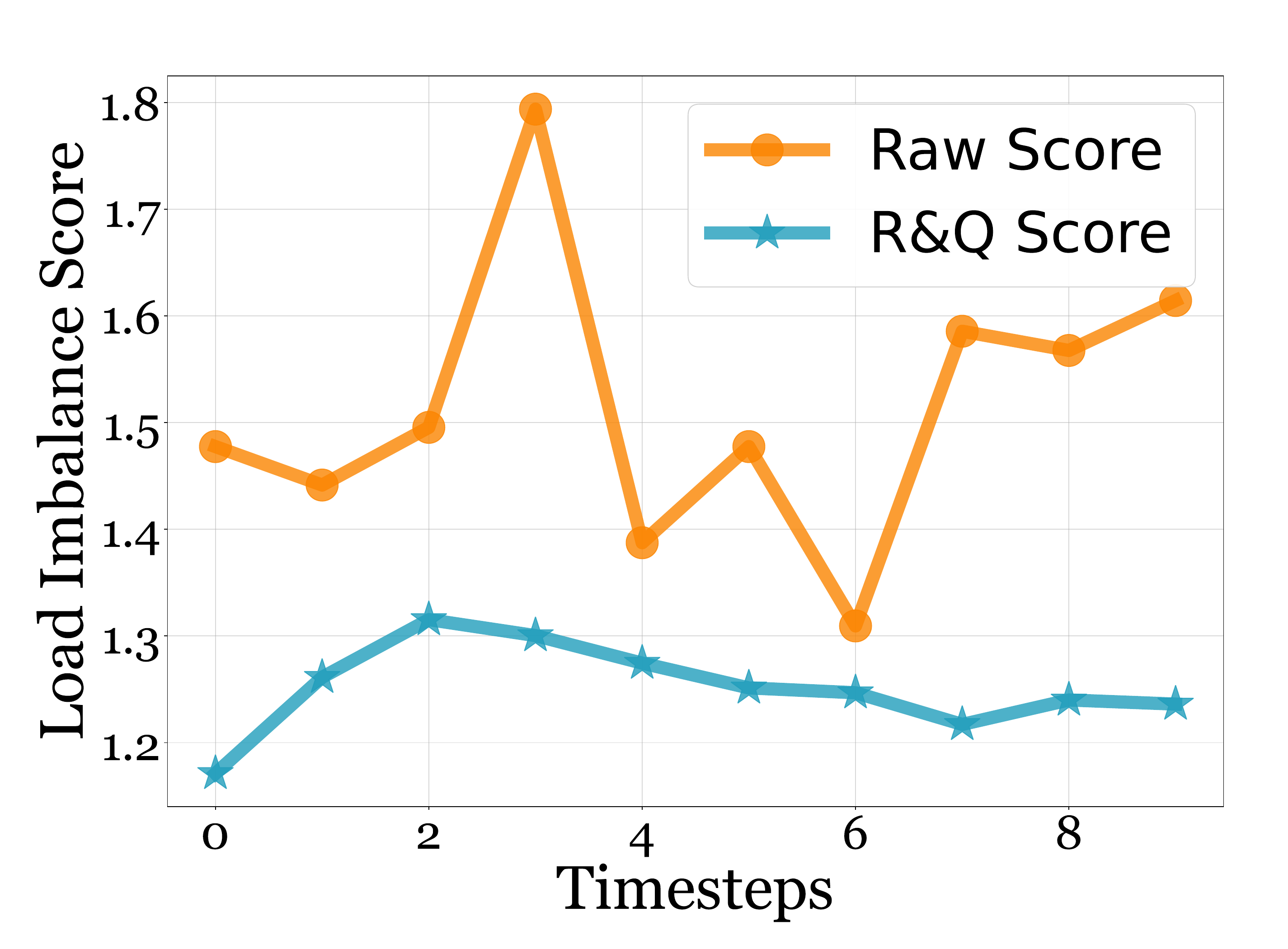}
        \caption{Layer 9}
    \end{subfigure}
    \hfill
    \begin{subfigure}{0.24\textwidth}
        \centering
        \includegraphics[width=\linewidth]{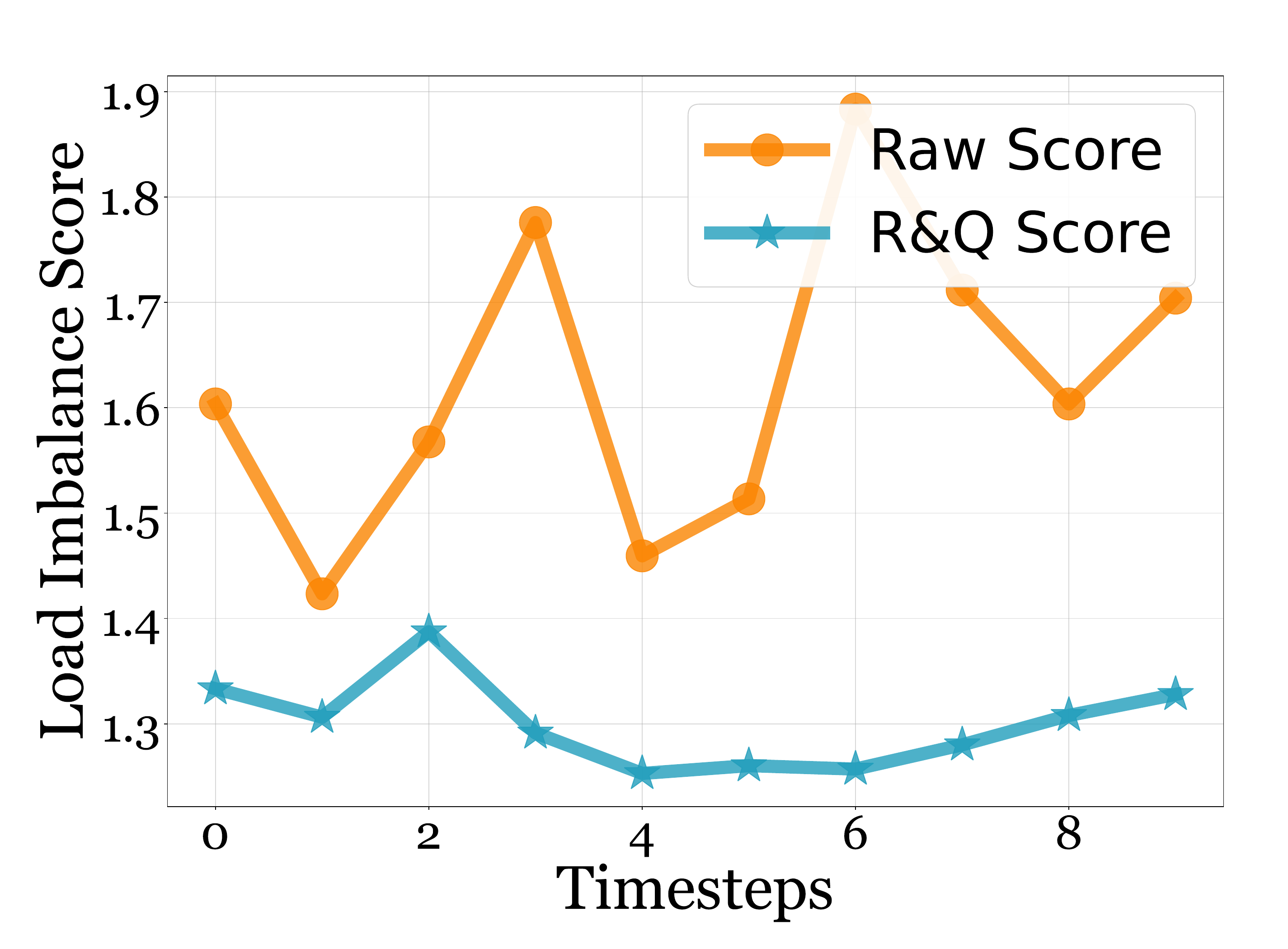}
        \caption{Layer 11}
    \end{subfigure}

    \begin{subfigure}{0.24\textwidth}
        \centering
        \includegraphics[width=\linewidth]{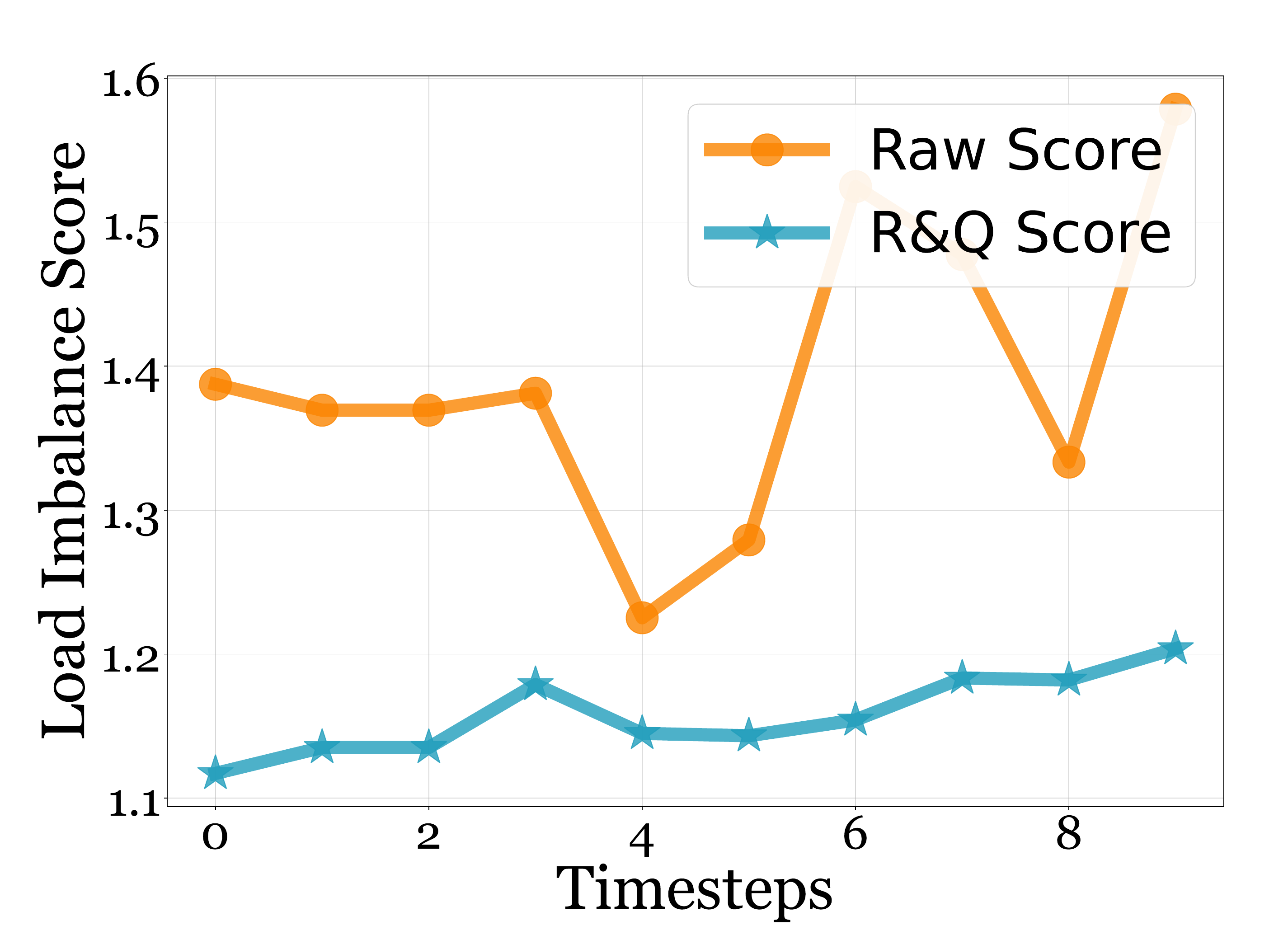}
        \caption{Layer 16}
    \end{subfigure}
    \hfill
    \begin{subfigure}{0.24\textwidth}
        \centering
        \includegraphics[width=\linewidth]{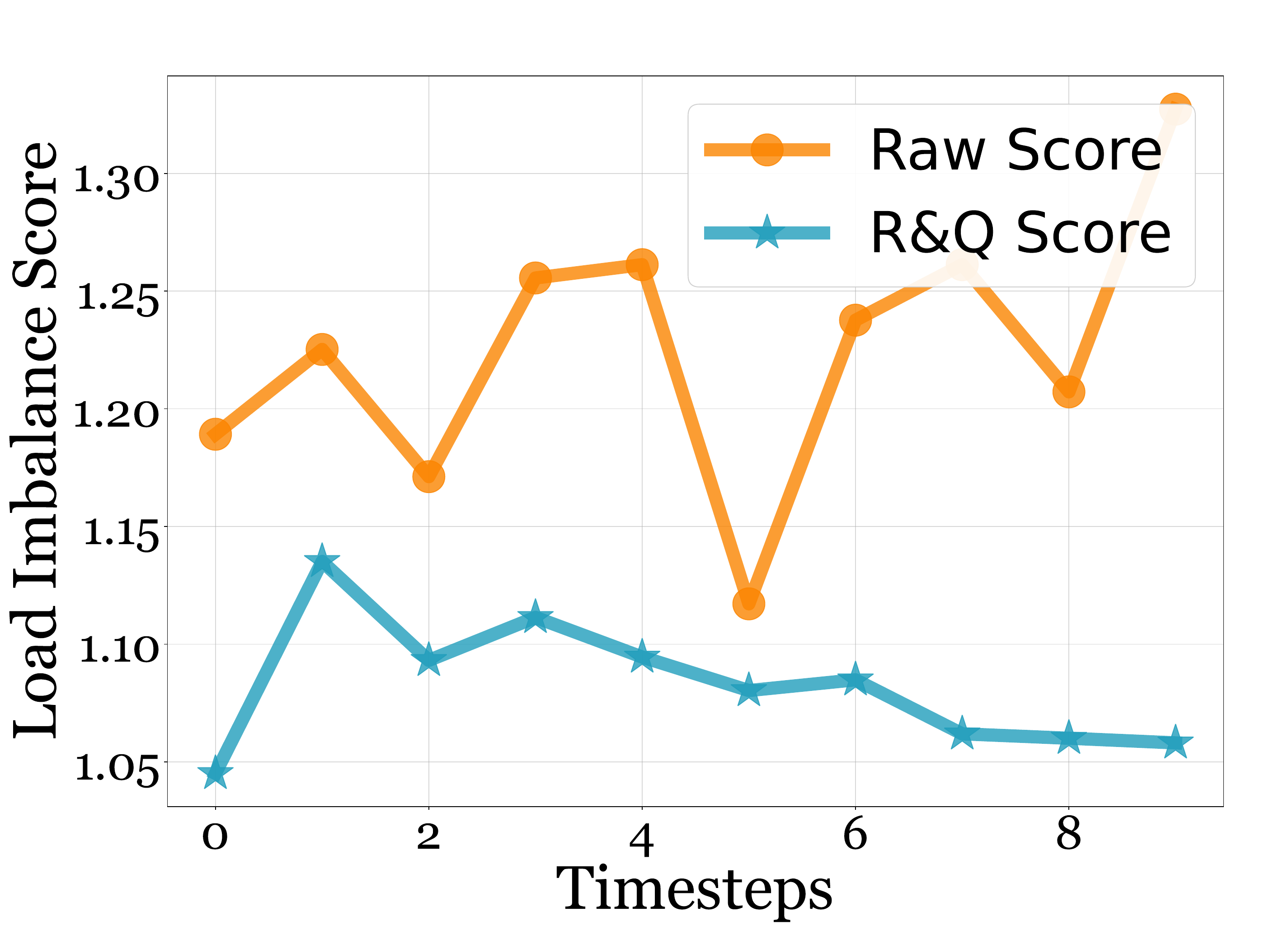}
        \caption{Layer 21}
    \end{subfigure}
    \hfill
    \begin{subfigure}{0.24\textwidth}
        \centering
        \includegraphics[width=\linewidth]{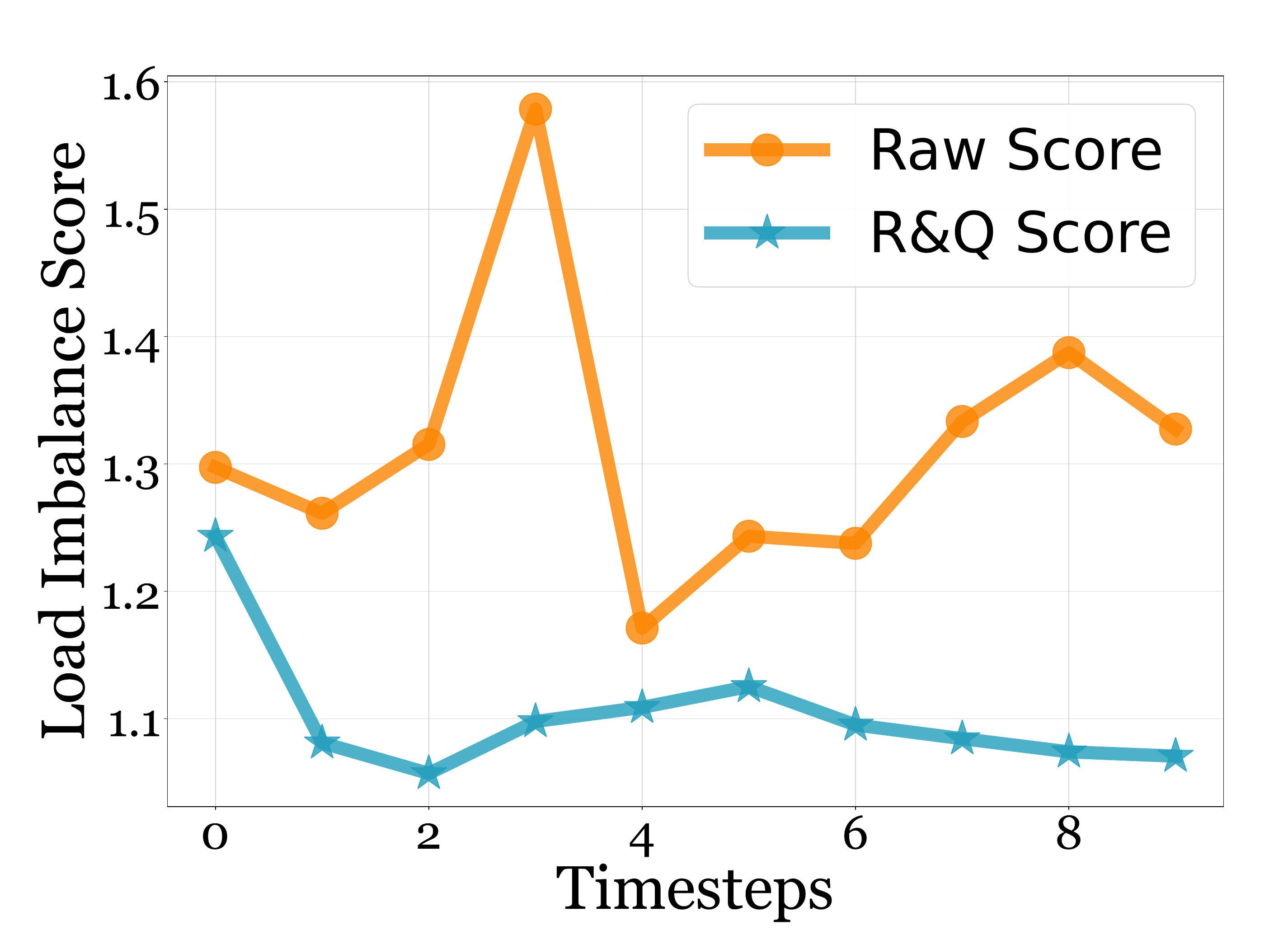}
        \caption{Layer 24}
    \end{subfigure}
    \hfill
    \begin{subfigure}{0.24\textwidth}
        \centering
        \includegraphics[width=\linewidth]{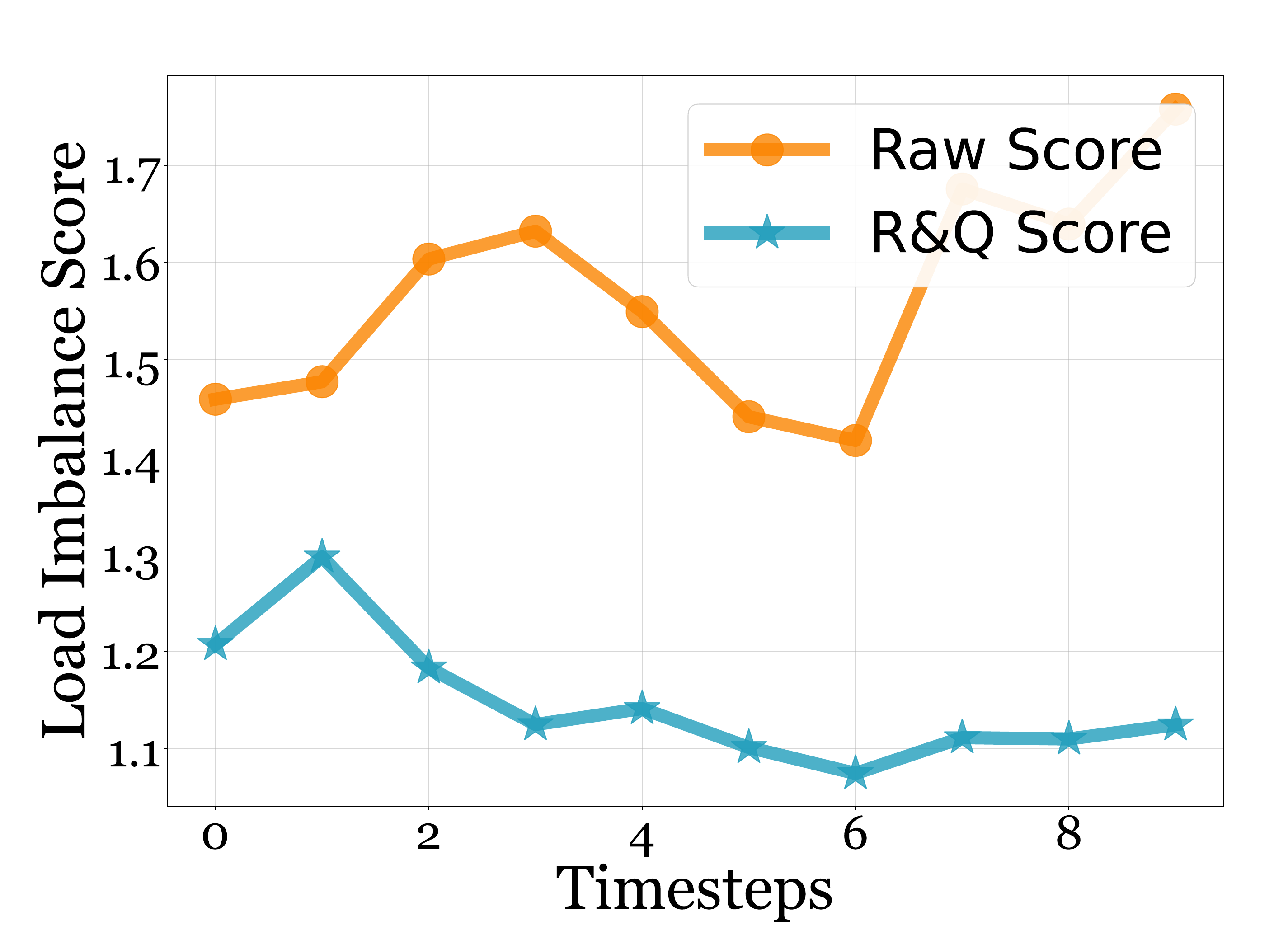}
        \caption{Layer 30}
    \end{subfigure}
    \vspace{-3mm}
    
    \vskip\baselineskip
    \begin{subfigure}{0.24\textwidth}
        \centering
        \includegraphics[width=\linewidth]{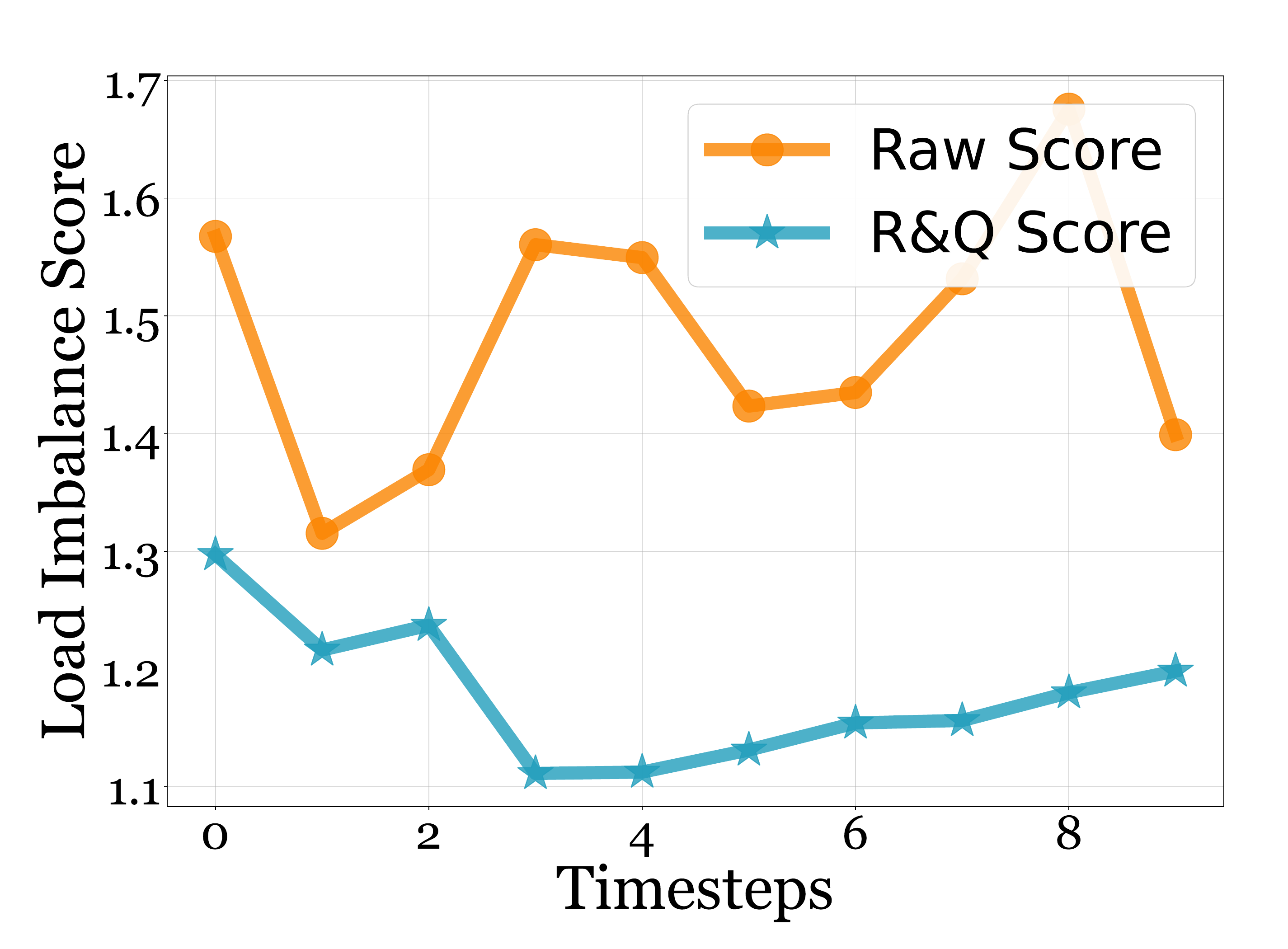}
        \caption{Layer 1}
    \end{subfigure}
    \hfill
    \begin{subfigure}{0.24\textwidth}
        \centering
        \includegraphics[width=\linewidth]{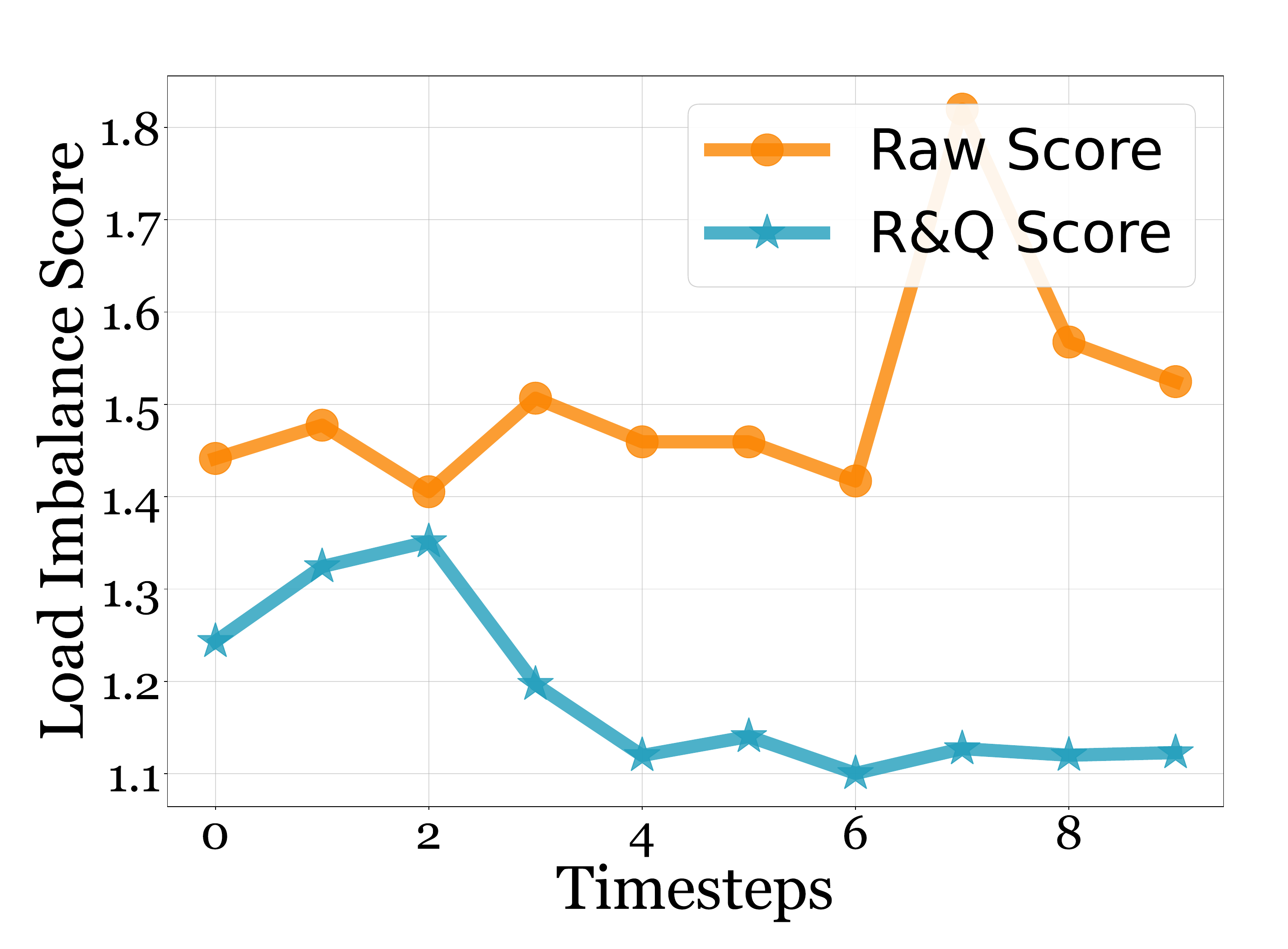}
        \caption{Layer 4}
    \end{subfigure}
    \hfill
    \begin{subfigure}{0.24\textwidth}
        \centering
        \includegraphics[width=\linewidth]{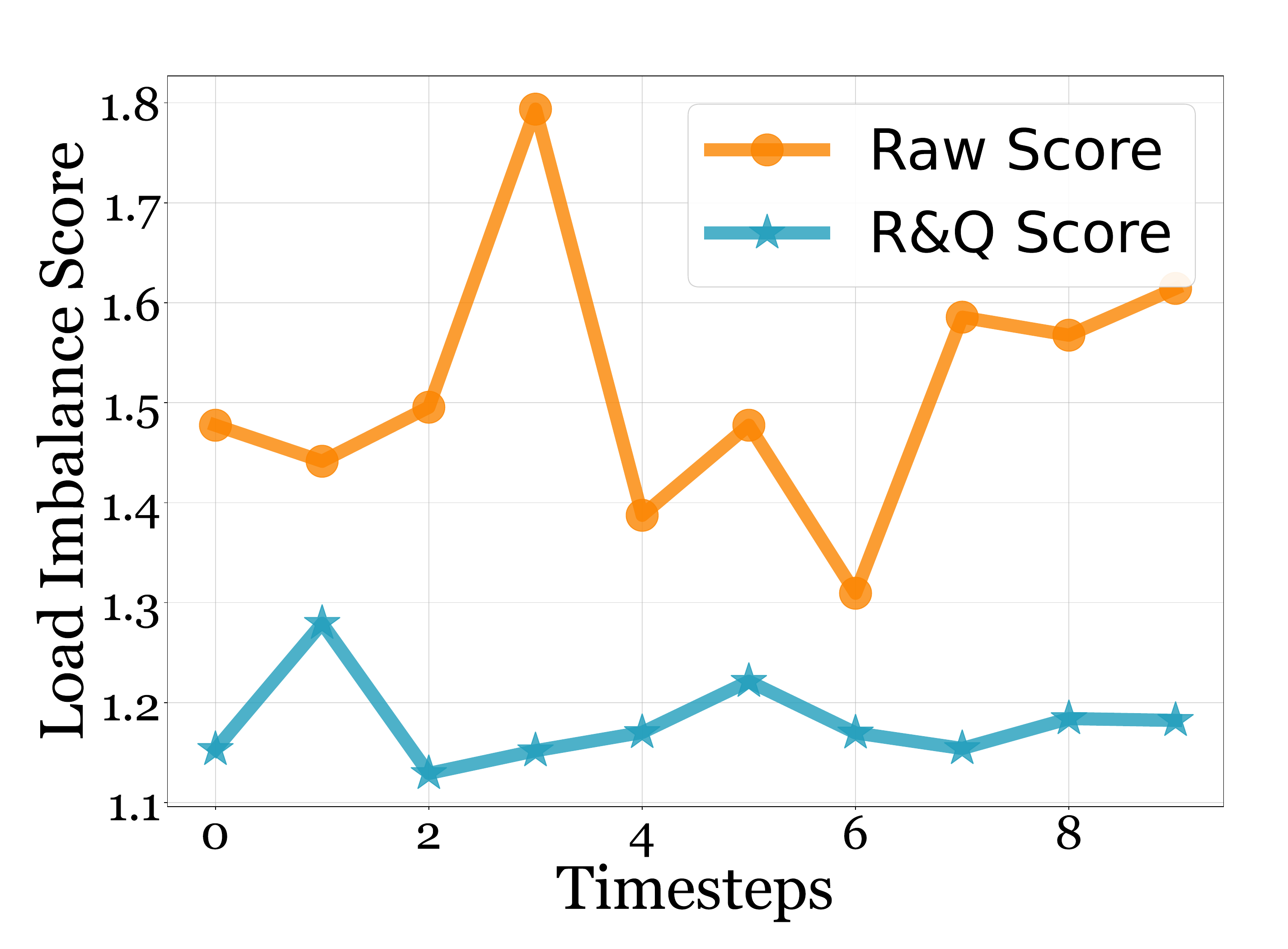}
        \caption{Layer 9}
    \end{subfigure}
    \hfill
    \begin{subfigure}{0.24\textwidth}
        \centering
        \includegraphics[width=\linewidth]{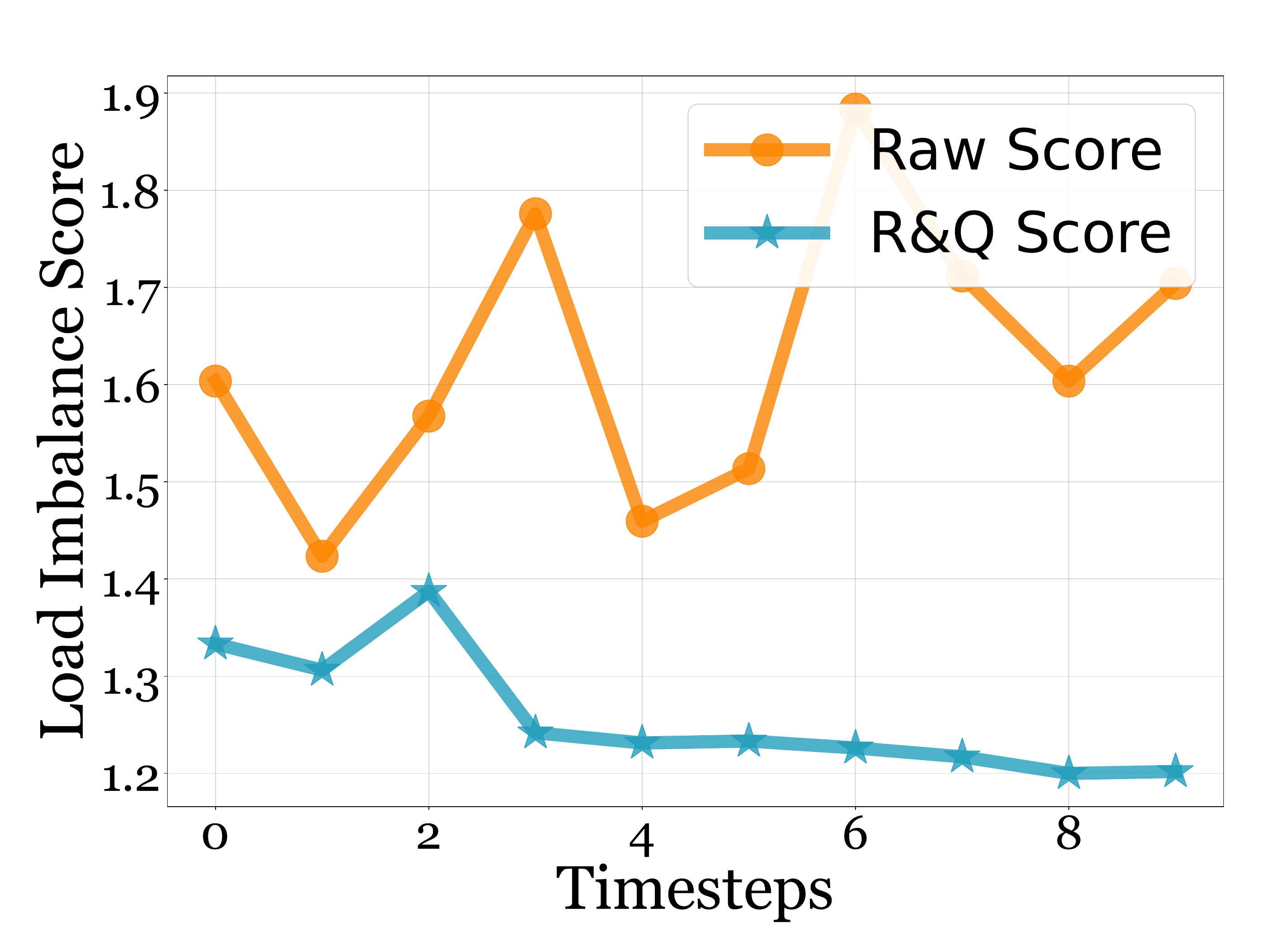}
        \caption{Layer 11}
    \end{subfigure}
    \vspace{-3mm}

    \vskip\baselineskip
    \begin{subfigure}{0.24\textwidth}
        \centering
        \includegraphics[width=\linewidth]{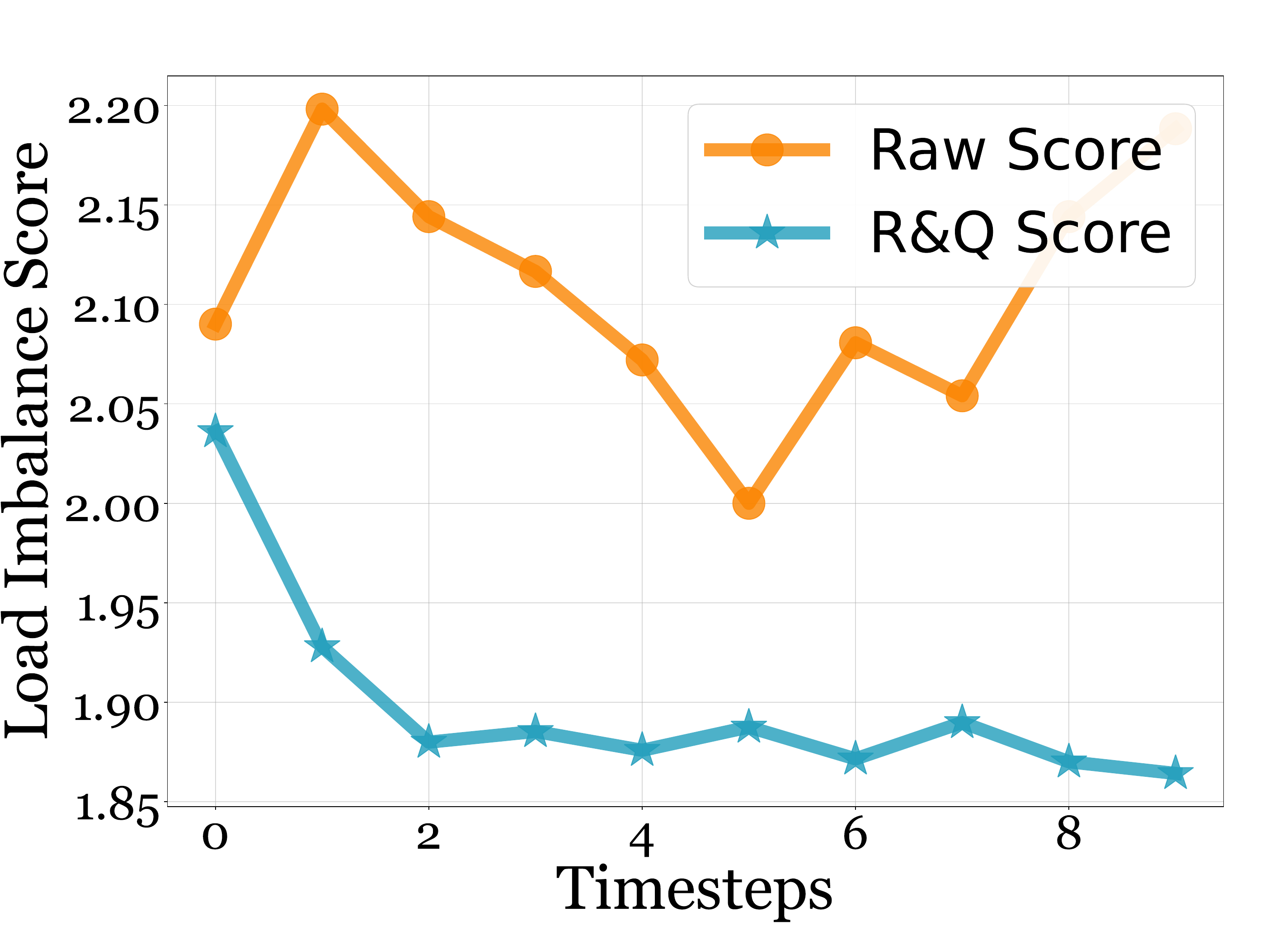}
        \caption{Layer 16}
    \end{subfigure}
    \hfill
    \begin{subfigure}{0.24\textwidth}
        \centering
        \includegraphics[width=\linewidth]{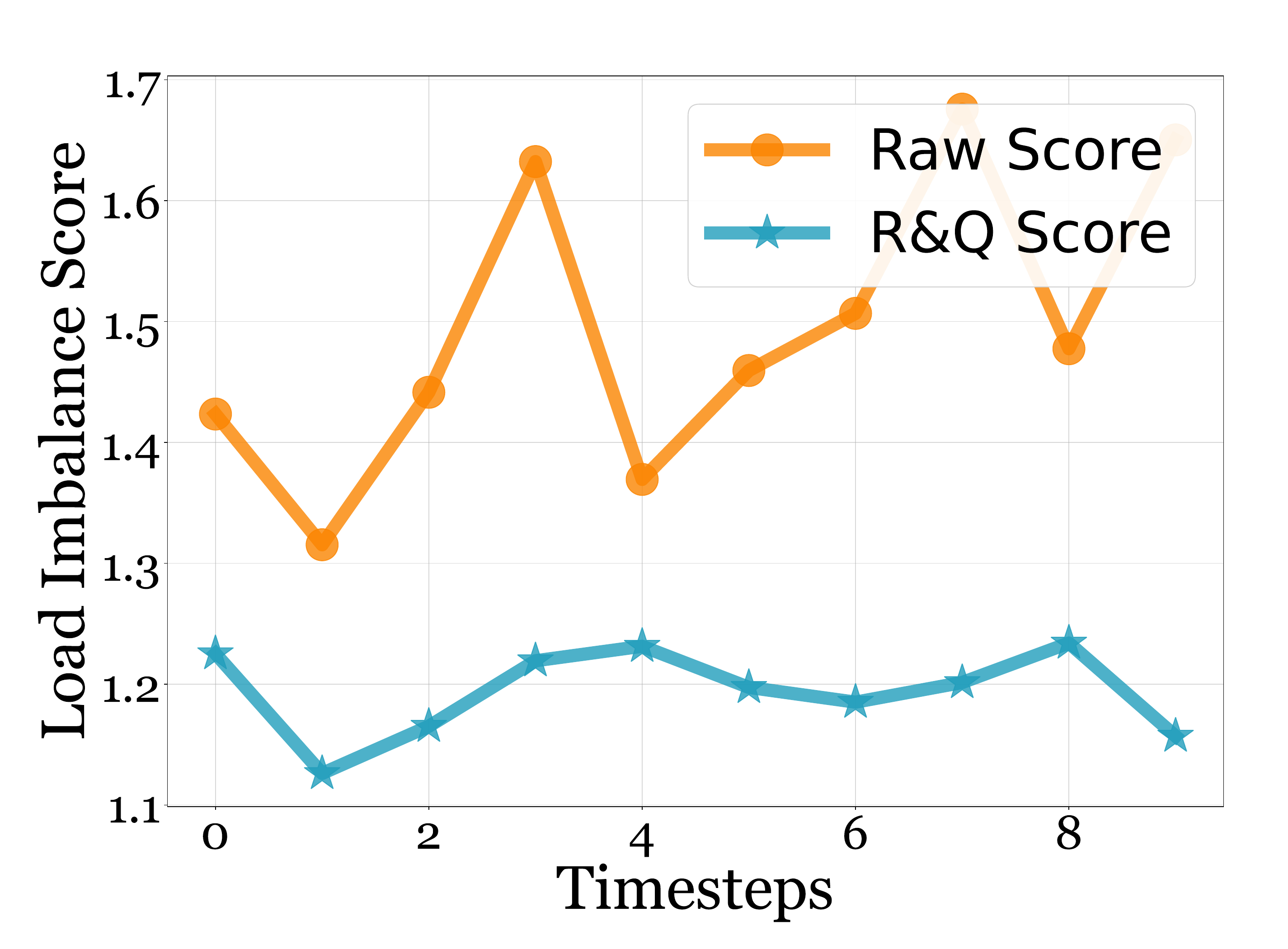}
        \caption{Layer 21}
    \end{subfigure}
    \hfill
    \begin{subfigure}{0.24\textwidth}
        \centering
        \includegraphics[width=\linewidth]{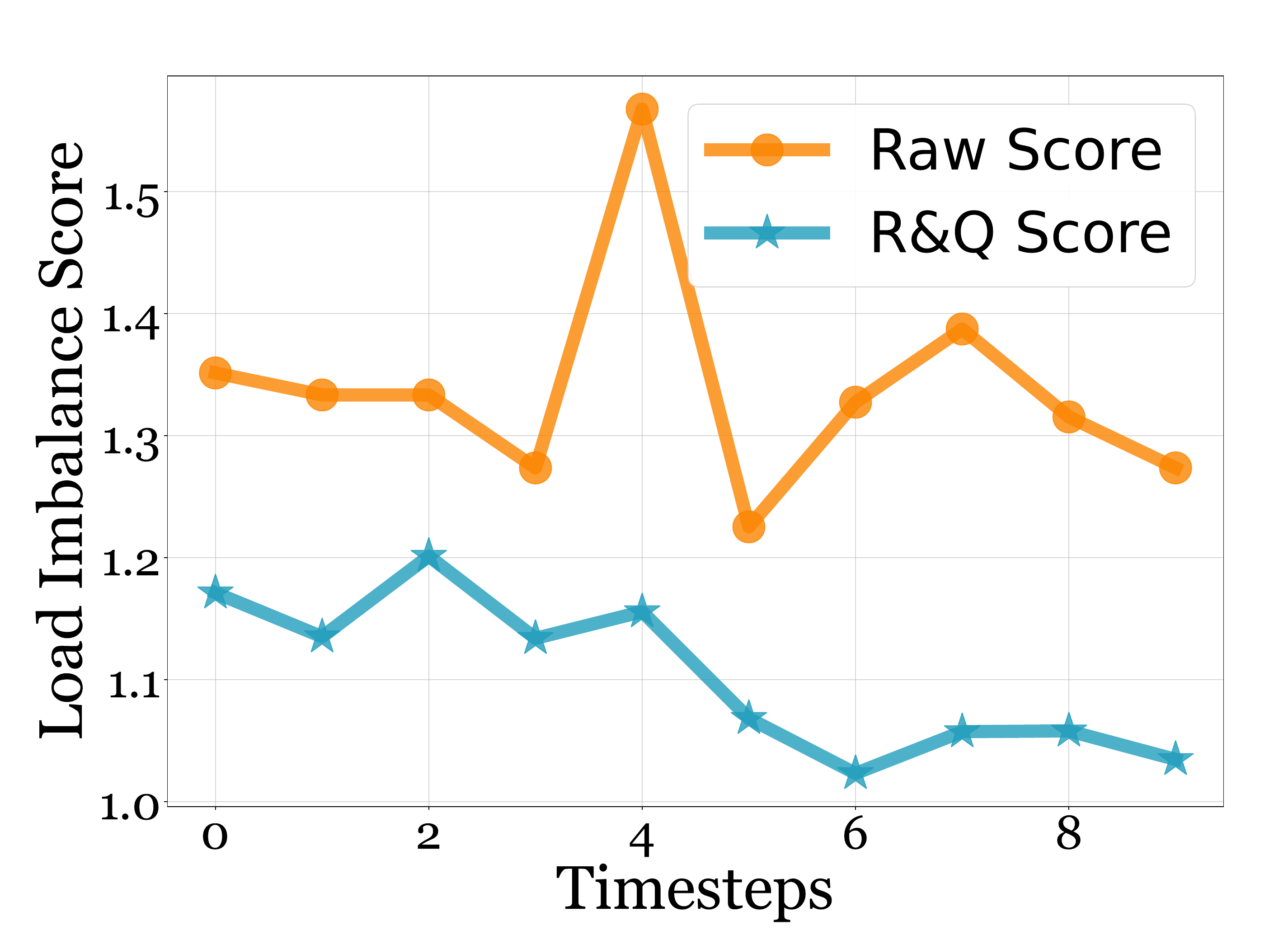}
        \caption{Layer 24}
    \end{subfigure}
    \hfill
    \begin{subfigure}{0.24\textwidth}
        \centering
        \includegraphics[width=\linewidth]{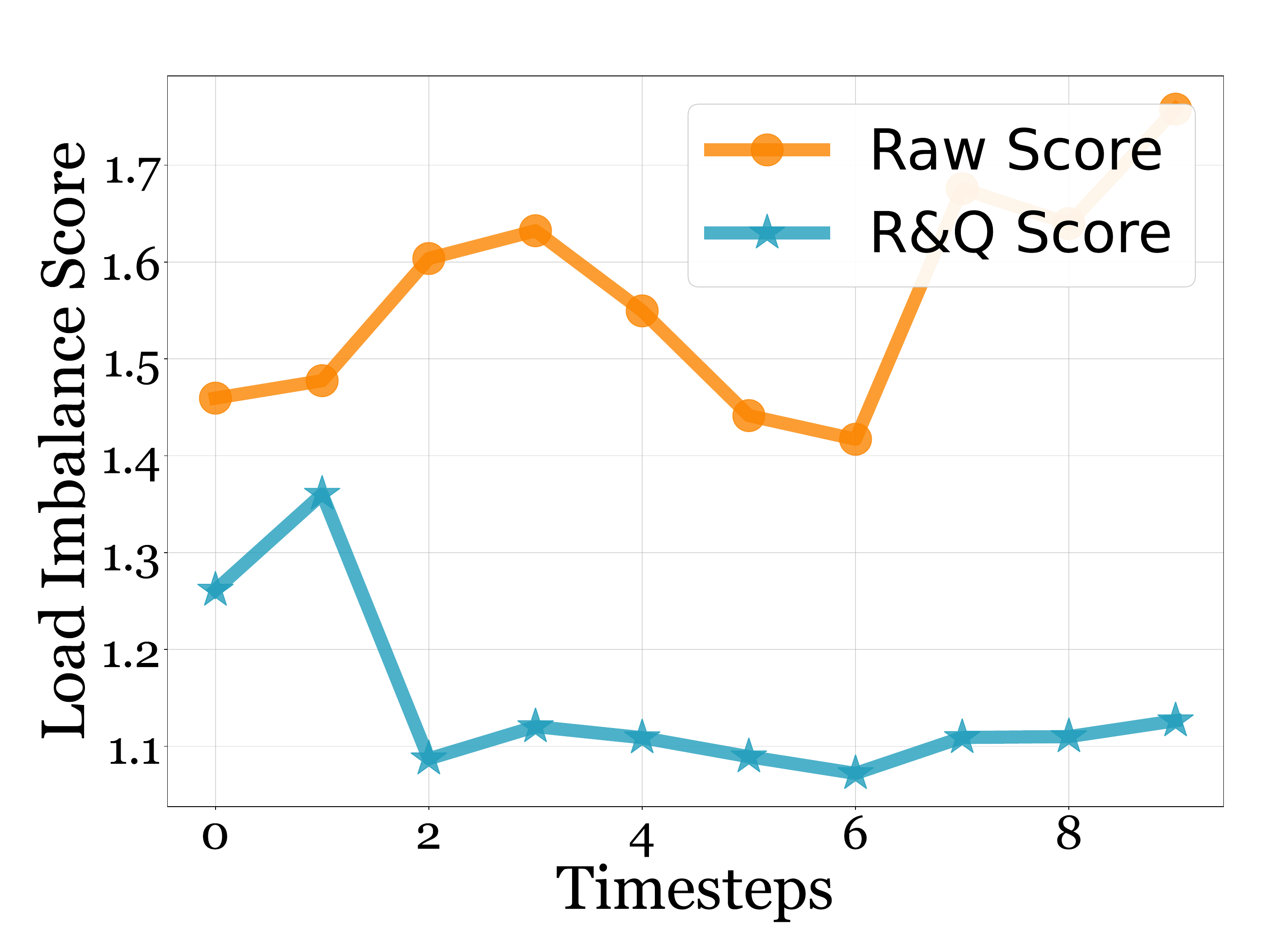}
        \caption{Layer 30}
    \end{subfigure}

    \caption{
        \textbf{Temporal Load Imbalance Comparison under Streaming Inputs.}
        This figure compares the LIS (Definition~\ref{def:lb_score}) between the baseline (Raw) and the proposed Replicate-and-Quantize (R\&Q) framework on the LLaMA-MoE model during streaming inference.
        Each subplot reports LIS over timesteps for a representative MoE layer.
        The top two rows depict cumulative LIS computed over all prior timesteps, reflecting long-term balancing stability, 
        while the bottom two rows show instantaneous LIS based on the immediately preceding timestep, revealing short-term fluctuations.
        Across all layers, R\&Q consistently reduces both cumulative and instantaneous imbalance, demonstrating improved routing stability over time.
        }
    \label{fig:streaming_data}
\end{figure*}

\subsection{Results}

In this section, we want to validate the effectiveness of our proposed replicate-and-quantize strategy. We would like to answer the following research questions:

\begin{table*}[ht!]
\centering
\small
\setlength{\tabcolsep}{5pt}
\renewcommand{\arraystretch}{1.12}
\adjustbox{width=0.90\textwidth}{%
\begin{tabular}{
@{}c
@{\hspace{8pt}}l
@{\hspace{8pt}}r
@{\hspace{8pt}}c  
@{\hspace{8pt}}c     
@{\hspace{20pt}}l     
@{\hspace{8pt}}l
@{\hspace{8pt}}r@{}}
\hline
\multirow{2}{*}{Model} & \multirow{2}{*}{Dataset} 
& \multicolumn{2}{c}{LIS} & \multirow{2}{*}{\(\Delta\)} 
& \multicolumn{2}{c}{Accuracy (\%)} & \multirow{2}{*}{\(\Delta\)} \\ 
\cmidrule(lr){3-4} \cmidrule(lr){6-7}
 & & Raw & R\&Q &  & Raw & R\&Q &  \\ \midrule
\multirow{6}{*}{\makecell{Switch Transformer \\ (8 experts)}}  
& GSM8K        & 1.9709 & \textbf{1.3937} & +0.5772 & \textbackslash{} & \textbackslash{} & -- \\
& Truthful QA  & 1.4956 & \textbf{1.3494} & +0.1462 & \textbf{36.9\% ± 1.1\%} & 36.4\% ± 1.1\% & -0.5\% \\
& Winogrande   & 1.5261 & \textbf{1.2146} & +0.3115 & \textbf{49.6\% ± 1.4\%} & 49.2\% ± 1.4\% & -0.4\% \\
& Hellaswag    & 1.4182 & \textbf{1.3623} & +0.0559 & 27.5\% ± 0.5\% & \textbf{27.6\% ± 0.5\%} & +0.1\% \\
& MMLU         & 1.5405 & \textbf{1.2962} & +0.2443 & 23.0\% ± 0.4\% & \textbf{25.2\% ± 0.4\%} & +2.2\% \\
& PIQA         & 1.5770 & \textbf{1.2756} & +0.3014 & 58.1\% ± 1.2\% & \textbf{58.3\% ± 1.2\%} & +0.2\% \\ \midrule

\multirow{6}{*}{\makecell{Switch Transformer \\ (16 experts)}} 
& GSM8K        & 2.6937 & \textbf{1.9582} & +0.7355 & \textbackslash{} & \textbackslash{} & -- \\
& Truthful QA  & 2.0121 & \textbf{1.6856} & +0.3265 & \textbf{37.4\% ± 1.1\%} & 36.9\% ± 1.1\% & -0.5\% \\
& Winogrande   & 2.0063 & \textbf{1.5394} & +0.4669 & \textbf{49.0\% ± 1.4\%} & 48.5\% ± 1.4\% & -0.5\% \\
& Hellaswag    & 1.9681 & \textbf{1.7369} & +0.2312 & 28.6\% ± 0.5\% & \textbf{28.7\% ± 0.5\%} & +0.1\% \\
& MMLU         & 1.9964 & \textbf{1.7067} & +0.2897 & 23.0\% ± 0.4\% & \textbf{25.0\% ± 0.4\%} & +2.0\% \\
& PIQA         & 2.0355 & \textbf{1.6080} & +0.4275 & 54.6\% ± 1.2\% & \textbf{55.0\% ± 1.2\%} & +0.4\% \\ \midrule

\multirow{6}{*}{LLaMa-MoE}                       
& GSM8K        & 1.3565 & \textbf{1.1963} & +0.1602 & 3.5\% ± 0.5\% & \textbf{3.6\% ± 0.5\%} & +0.1\% \\
& Truthful QA  & 1.2946 & \textbf{1.2025} & +0.0921 & 27.3\% ± 1.0\% & \textbf{27.3\% ± 1.0\%} & 0.0\% \\
& Winogrande   & 1.3925 & \textbf{1.2791} & +0.1134 & \textbf{67.3\% ± 1.3\%} & 66.7\% ± 1.3\% & -0.6\% \\
& Hellaswag    & 1.3047 & \textbf{1.2258} & +0.0789 & \textbf{54.1\% ± 0.5\%} & 54.0\% ± 0.5\% & -0.1\% \\
& MMLU         & 1.3289 & \textbf{1.1964} & +0.1325 & 28.0\% ± 0.4\% & \textbf{28.0\% ± 0.4\%} & 0.0\% \\
& PIQA         & 1.2943 & \textbf{1.2366} & +0.0577 & 77.0\% ± 1.0\% & \textbf{77.0\% ± 1.0\%} & 0.0\% \\ \midrule

\multirow{6}{*}{DeepSeek-MoE}                    
& GSM8K        & 4.8182 & \textbf{3.9725} & +0.8457 & \textbf{15.6\% ± 1.0\%} & 15.4\% ± 1.0\% & -0.2\% \\
& Truthful QA  & 2.9672 & \textbf{2.1749} & +0.7923 & \textbf{31.1\% ± 1.0\%} & 31.1\% ± 1.0\% & 0.0\% \\
& Winogrande   & 3.1386 & \textbf{2.4475} & +0.6911 & 70.0\% ± 1.3\% & \textbf{70.6\% ± 1.3\%} & +0.6\% \\
& Hellaswag    & 3.0387 & \textbf{2.6403} & +0.3984 & \textbf{59.8\% ± 0.5\%} & 59.8\% ± 0.5\% & 0.0\% \\
& MMLU         & 3.7359 & \textbf{2.8222} & +0.9137 & 44.7\% ± 0.4\% & \textbf{44.7\% ± 0.4\%} & 0.0\% \\
& PIQA         & 4.0780 & \textbf{3.0585} & +1.0195 & 78.8\% ± 1.0\% & \textbf{79.1\% ± 1.0\%} & +0.3\% \\ \midrule

\multirow{6}{*}{DeepSeek V2 Lite}                
& GSM8K        & 4.7886 & \textbf{3.5998} & +1.1888 & \textbf{38.4\% ± 1.3\%} & 37.2\% ± 1.3\% & -1.2\% \\
& Truthful QA  & 3.1838 & \textbf{2.3862} & +0.7976 & 34.9\% ± 1.7\% & \textbf{35.0\% ± 1.7\%} & +0.1\% \\
& Winogrande   & 3.4756 & \textbf{2.7207} & +0.7549 & 70.8\% ± 1.3\% & \textbf{71.4\% ± 1.3\%} & +0.6\% \\
& Hellaswag    & 3.1611 & \textbf{2.9341} & +0.2270 & 58.8\% ± 0.5\% & \textbf{60.4\% ± 0.5\%} & +1.6\% \\
& MMLU         & 4.0509 & \textbf{3.0119} & +1.0390 & 55.0\% ± 0.4\% & \textbf{58.0\% ± 0.4\%} & +3.0\% \\
& PIQA         & 4.3504 & \textbf{3.2925} & +1.0579 & 79.8\% ± 0.9\% & \textbf{79.9\% ± 0.9\%} & +0.1\% \\ \bottomrule
\end{tabular}%
}
\caption{\textbf{Main evaluation of the Replicate-and-Quantize (R\&Q) strategy across different SMoE architectures.}
We report both LIS - lower is better, and task accuracy (\%).  
Across all evaluated models, R\&Q substantially reduces LIS (average reduction: 0.5–1.0) while maintaining or slightly improving accuracy.  
This demonstrates that R\&Q effectively redistributes expert utilization and mitigates routing imbalance without retraining or accuracy degradation.  
Results on GSM8K for the Switch Transformer are omitted as the model fails to converge on this mathematical reasoning task.
}
\label{tab:Table1}
\end{table*}

To evaluate the effectiveness of our proposed Replicate-and-Quantize (R\&Q) strategy, we analyze expert token routing patterns in the Switch Transformer. Figure~\ref{fig:combined} visualize the expert activation gap—defined as the token count difference between consecutively ranked experts—across layers. In the baseline model (Table~\ref{tab:Table1}), token assignments are strongly imbalanced, with a few experts disproportionately activated, leading to pronounced load imbalance. In contrast, R\&Q (Table~\ref{tab:Table1}) significantly reduces these disparities, resulting in a more even and uniform token distribution across experts. This visual evidence confirms that R\&Q mitigates expert overload by redistributing workload without modifying the routing mechanism, ultimately enabling more balanced and efficient inference-time execution.
\begin{itemize}
    \item \textbf{Option 1 (Figure~\ref{fig:streaming_data} (Top two rows))}: At each timestep, our method determines which expert needs to be replicated based on \textit{all previous timesteps}. Quantized experts are pre-identified using 10\% of the MMLU dataset.
    \item \textbf{Option 2 (Figure~\ref{fig:streaming_data} (Bottom two rows))}: The decision to replicate an expert at the current timestep relies solely on the statistics from the \textit{immediately previous} timestep.
\end{itemize}

As shown in Figure~\ref{fig:streaming_data}, our proposed R\&Q method consistently achieves significantly lower LIS across all timesteps, compared to the baseline Raw model. This indicates that our method successfully alleviates the load imbalance problem, even under adversarial input conditions. Notably, both strategy variants—whether relying on full historical data (Option 1) or only the most recent window (Option 2)—lead to clear and stable reductions in load imbalance over time.

\noindent \textbf{Q1: Does Replicate-and-Quantize (R\&Q) effectively improve load balance in pre-trained SMoEs? A1:R\&Q consistently improves expert load balance across all evaluated tasks.}
We evaluate the impact of the proposed R\&Q strategy on four representative pre-trained Sparse Mixture-of-Experts (SMoE) models—Switch Transformer, LLaMa-MoE, DeepSeek-MoE, and DeepSeek V2 Lite—across six diverse downstream tasks. As summarized in Table~\ref{tab:Table1}, R\&Q consistently reduces expert load imbalance across all settings, confirming its robustness as a post-hoc adaptation method. For example, in the Switch Transformer with eight experts, the LIS on GSM8K drops from 1.9709 to 1.3937, while DeepSeek V2 Lite on PIQA achieves a similar improvement from 4.3504 to 3.2925. These consistent reductions indicate that R\&Q effectively redistributes token-level computation, mitigating routing skew and promoting more uniform expert utilization.

Importantly, the magnitude of improvement correlates with the initial imbalance severity. Models exhibiting the most skewed token distributions (e.g., DeepSeek V2 Lite) experience the largest absolute LIS reductions—exceeding 1.0 on datasets such as MMLU and GSM8K—while more balanced architectures like LLaMa-MoE still benefit from moderate but stable improvements (e.g., from 1.3565 to 1.1963 on GSM8K). This suggests that R\&Q adapts proportionally to each model’s internal routing dynamics, automatically focusing its replication effort on overburdened experts without introducing redundancy in balanced layers.

Beyond numerical gains, these results provide empirical evidence for R\&Q’s dual mechanism: (\textit{i}) expert replication distributes incoming token traffic, reducing localized computational bottlenecks, and (\textit{ii}) quantization of less important experts ensures that the resulting duplication does not violate memory or latency budgets. Because both steps are applied \textit{post-training} and require no retraining or router modification, R\&Q serves as a lightweight, architecture-agnostic remedy for deployment-time imbalance. Overall, these findings underscore that R\&Q not only improves efficiency but also enhances fairness in expert activation, a property critical for scaling MoE systems to large and heterogeneous workloads.

\noindent \textbf{Q2: Can the R\&Q strategy maintain or improve predictive performance in SMoEs?
A2: Preserved or enhanced model accuracy}
In addition to improving load balance, the proposed R\&Q strategy preserves—and in certain cases, slightly enhances—the predictive performance of pre-trained SMoE models. As shown in the accuracy columns of Table~\ref{tab:Table1}, performance differences between the raw and R\&Q-processed models are marginal, often within the standard deviation of repeated runs. Notably, several benchmarks exhibit consistent accuracy gains: on the MMLU task, the Switch Transformer with 16 experts improves from 23.0\% to 25.0\%, while DeepSeek V2 Lite improves from 55\% to 58\%. These improvements indicate that R\&Q does not merely preserve accuracy but may also facilitate better generalization and calibration in downstream evaluation.

A closer inspection reveals that this stability arises from R\&Q’s architecture-preserving design. Because the strategy operates entirely at the inference level—without altering routing logic or retraining the model—it avoids introducing distributional shifts that typically degrade post-hoc modified MoE systems. The replication step mitigates local saturation by redistributing input tokens across duplicated experts, effectively reducing gradient interference and expert competition observed during routing. Meanwhile, precision-aware quantization selectively compresses less important experts, maintaining representational fidelity where it matters most. This asymmetric treatment of experts helps preserve both expressivity and task-level robustness, explaining the small yet consistent performance improvements observed across tasks.

\begin{wrapfigure}{r}{0.52\textwidth}
\vspace{-10pt}

\begin{minipage}{0.50\textwidth}
\centering
\small

\setlength{\tabcolsep}{6pt}
\renewcommand{\arraystretch}{1.05}

\resizebox{\textwidth}{!}{
\begin{tabular}{@{}lcccc@{}}
    \toprule
    \textbf{Dataset} & \multicolumn{2}{c}{\textbf{Accuracy}} & \multicolumn{2}{c}{\textbf{LIS}} \\ 
    \cmidrule(lr){2-3} \cmidrule(lr){4-5}
     & R\&Q & Raw & R\&Q & Raw \\ 
    \midrule
    codexglue(code2text) & 1.5517 & 1.5463 & 1.9426 & 2.6119 \\
    coqa                 & 0.0104 & 0.0106 & 1.6488 & 2.0412 \\
    wikitext             & 17.2096 & 17.2826 & 1.6235 & 2.0900 \\ 
    \bottomrule
\end{tabular}
}

\captionsetup{type=table}
\caption{\textbf{Performance on generation benchmarks using Switch Transformer (8 experts).}
This table reports accuracy and LIS across three representative generation tasks. 
While R\&Q maintains accuracy nearly identical to the unmodified (Raw) model, it consistently achieves substantially lower imbalance scores, confirming its ability to improve inference-time expert utilization without harming task performance.}

\label{tab:challenge}

\end{minipage}
\vspace{-8pt}
\end{wrapfigure}

Overall, Table~\ref{tab:Table1} demonstrates that R\&Q successfully addresses the core challenge of expert load imbalance while maintaining or slightly enhancing the predictive quality of SMoE models. These findings strongly support both RQ1 and RQ2: R\&Q enables more balanced computation and more reliable inference without sacrificing accuracy, thereby offering a practical and deployment-ready solution for large-scale MoE architectures.

\noindent \textbf{Q3: Does R\&Q maintain effective load balance under streaming data scenarios?
A3: R\&Q preserves stable and balanced expert utilization even with continuously incoming data.}
To evaluate load-balancing stability under non-stationary inputs, we simulate a streaming inference scenario using the MMLU benchmark. Specifically, we first feed all tokens through the pretrained LLaMA-MoE model and record the number of tokens routed to each expert in every layer. Experts that receive disproportionately many tokens are identified as heavy-hitter experts, and the corresponding tokens are used to construct an adversarial streaming dataset. The data stream is divided into ten sequential segments, referred to as timesteps, each representing one batch of incoming data.

We examine two strategies for applying expert replication during inference:  
(i) using cumulative routing statistics from all previous timesteps (long-term strategy), and  
(ii) using statistics from only the most recent window (short-term strategy).  

Figure~\ref{fig:streaming_data} compares the temporal LIS (Definition~\ref{def:lb_score}) between the baseline model and our R\&Q framework. The upper two rows report cumulative LIS computed over all prior timesteps, reflecting long-term balancing stability, while the lower two rows show instantaneous LIS based on the immediately preceding timestep, capturing short-term fluctuations. Across all representative MoE layers, R\&Q consistently achieves substantially lower LIS values than the baseline, indicating reduced load imbalance and improved routing stability over time. Notably, both replication strategies exhibit comparable effectiveness, demonstrating that R\&Q robustly maintains balance under streaming conditions without retraining or additional overhead.

\begin{wrapfigure}{r}{0.5\textwidth}
\vspace{-10pt}

\centering
\includegraphics[width=0.48\textwidth]{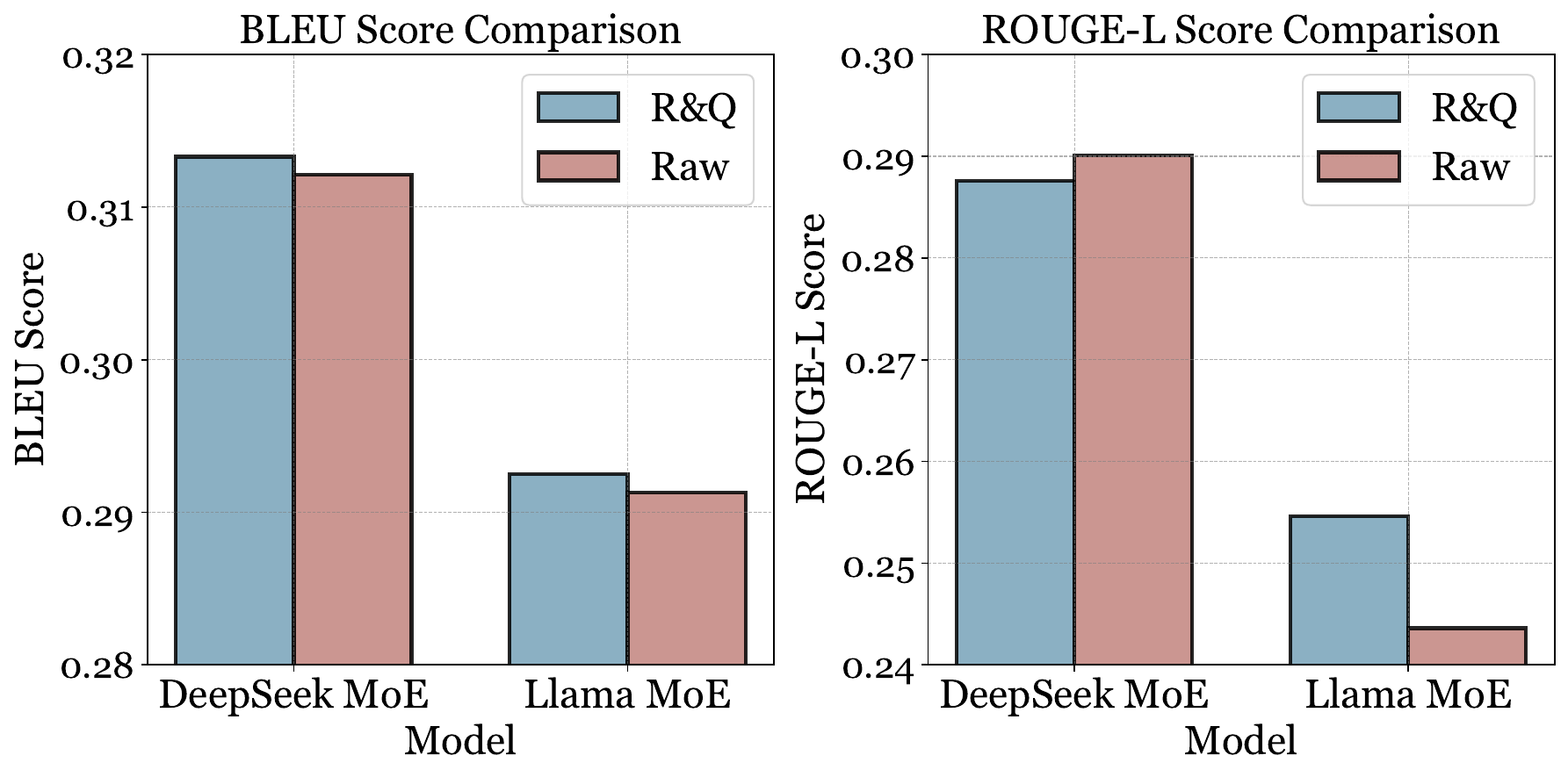}

\caption{\textbf{BLEU and ROUGE-L Evaluation on TruthfulQA Across Models.} 
This figure compares BLEU and ROUGE-L scores of DeepSeek-MoE and LLaMA-MoE models under R\&Q and raw settings. 
The R\&Q configuration maintains comparable or slightly improved generation quality across both metrics, 
demonstrating that the proposed adaptation enhances load balance without degrading linguistic fidelity.}

\label{fig:bleu_rouge_comparison}
\vspace{-6pt}
\end{wrapfigure}

\noindent \textbf{Q4: How does R\&Q perform on challenging generation tasks? A4: Strong performance and robustness across diverse tasks.}  
We evaluate our proposed R\&Q strategy on a set of challenging generation tasks using the Switch Transformer (8 experts). The results, presented in Tables~\ref{tab:challenge} and Figure~\ref{fig:bleu_rouge_comparison}, include benchmarks such as CodexGLUE (code2text) \cite{lu2021codexgluemachinelearningbenchmark}, CoQA \cite{reddy-etal-2019-coqa}, and WikiText \cite{merity2016pointer}. We use task-appropriate evaluation metrics: \textit{smoothed BLEU} for CodexGLUE, \textit{F1 score} for CoQA, and \textit{byte-level perplexity} for WikiText. Additionally, we report LIS to assess the uniformity of expert activation.

Across all datasets, our method consistently demonstrates accuracy comparable to or surpassing the Raw baseline. Specifically, on CodexGLUE, it achieves a BLEU score of 1.5517, exceeding the baseline's 1.5463. On CoQA, it maintains a competitive F1 score of 0.0104, closely matching the baseline performance of 0.0106. Furthermore, the method improves language modeling quality on WikiText by reducing perplexity from 17.2826 to 17.2096. A notable advantage of the R\&Q strategy is its consistent ability to reduce expert load imbalance across various tasks. For instance, on the CodexGLUE benchmark, the imbalance score decreases substantially from 2.6119 to 1.9426. Comparable improvements are observed on CoQA and WikiText, where the scores drop from 2.0412 to 1.6488 and from 2.0900 to 1.6235, respectively. These reductions indicate a more even distribution of tokens across experts during inference, reflecting more efficient expert utilization and improved routing balance enabled by R\&Q.

We further validate the effectiveness of our strategy on the TruthfulQA benchmark by comparing BLEU and ROUGE-L scores across different MoE models (Figure~\ref{fig:bleu_rouge_comparison}). Both DeepSeek-MoE and LLaMA-MoE exhibit modest improvements with our method. For example, DeepSeek-MoE improves its BLEU score from 0.3121 to 0.3133, and LLaMA-MoE shows a ROUGE-L gain from 0.2436 to 0.2546. These results suggest that our approach supports more faithful and coherent generation without compromising computational efficiency. In summary, the R\&Q approach improves expert load balancing and maintains or slightly enhances task-specific performance across a variety of generation tasks and model architectures, demonstrating its practical value in large-scale MoE deployments.

\begin{table*}[t]
\centering
\footnotesize
\setlength{\tabcolsep}{6pt}
\renewcommand{\arraystretch}{1.08}
\resizebox{\textwidth}{!}{
\begin{tabular}{@{}lcccccc@{}}
\toprule
\textbf{Model} & \textbf{Method} & \textbf{Hellaswag} & \textbf{MMLU} & \textbf{PIQA} & \textbf{TruthfulQA} & \textbf{Winogrande} \\ 
\midrule

\multirow{4}{*}{\makecell[l]{Switch Transformer\\(8 Experts)}} 
 & Only Quantize Less-Important Experts & \textbf{27.95 ± 0.45} & 22.95 ± 0.35 & 57.51 ± 1.15 & 36.05 ± 1.10 & 51.38 ± 1.40 \\
 & Replicate Heavy-Hitter Experts + Quantize & 27.49 ± 0.45 & 24.98 ± 0.37 & 58.11 ± 1.15 & 36.35 ± 1.10 & 49.17 ± 1.41 \\
 & Quantize All Experts & 26.41 ± 0.44 & 22.95 ± 0.35 & 54.90 ± 1.16 & \textbf{37.75 ± 1.12} & \textbf{51.85 ± 1.40} \\
 & \textbf{R\&Q (Ours)} & 27.63 ± 0.45 & \textbf{25.22 ± 0.37} & \textbf{58.32 ± 1.15} & 36.40 ± 1.10 & 49.17 ± 1.41 \\ 
\midrule

\multirow{4}{*}{\makecell[l]{Switch Transformer\\(16 Experts)}} 
 & Only Quantize Less-Important Experts & 28.20 ± 0.45 & 22.95 ± 0.35 & 55.77 ± 1.16 & \textbf{39.14 ± 1.14} & 49.64 ± 1.41 \\
 & Replicate Heavy-Hitter Experts + Quantize & 28.64 ± 0.45 & 24.90 ± 0.36 & 54.90 ± 1.16 & 36.69 ± 1.12 & 48.30 ± 1.40 \\
 & Quantize All Experts & 25.95 ± 0.44 & 22.95 ± 0.35 & \textbf{55.82 ± 1.16} & 37.21 ± 1.10 & \textbf{50.91 ± 1.41} \\
 & \textbf{R\&Q (Ours)} & \textbf{28.72 ± 0.45} & \textbf{24.95 ± 0.36} & 55.01 ± 1.16 & 36.94 ± 1.12 & 48.54 ± 1.40 \\ 
\bottomrule
\end{tabular}
}
\caption{
\textbf{Ablation study of the Replicate-and-Quantize (R\&Q) framework on the Switch Transformer.}
We isolate the effects of (1) quantizing less-important experts, (2) replicating heavy-hitter experts, and (3) quantizing all experts.
All values denote accuracy (\%) averaged across multiple runs (\textit{mean ± std}).
R\&Q achieves the best or near-best performance across both 8- and 16-expert settings, demonstrating that combining replication and selective quantization yields the most effective trade-off between accuracy preservation and expert utilization.
}
\label{tab:abla}
\end{table*}

\subsection{Ablation Study}
To understand the contribution of each component in R\&Q, we conduct ablation experiments on both 8- and 16-expert Switch Transformers. Table~\ref{tab:abla} reports the results in terms of task accuracy. Quantizing only the less important experts achieves moderate accuracy while reducing model size, showing that lightweight compression alone can preserve much of the model’s capability. However, it does not mitigate inference-time imbalance, limiting efficiency gains. Replicating only a single heavy-hitter expert improves load distribution and inference stability, especially on compute-bound tasks such as MMLU and PIQA, but provides limited memory reduction. The full R\&Q configuration, which combines expert replication with selective quantization, consistently achieves the best overall accuracy—e.g., 58.3\% on PIQA (8 experts) and 55\% on the 16-expert setting—demonstrating the complementary nature of these two mechanisms.

\section{Related Works}
\paragraph{Efficiency and Load Balancing in Sparse Mixture-of-Experts}
The Mixture-of-Experts (MoE) framework, first introduced by ~\cite{jacobs1991adaptive}, decomposes complex problems into multiple sub-tasks handled by specialized experts. Its sparse variant (SMoE)~\cite{shazeer2017outrageously} activates only a subset of experts per token, achieving significant computational savings during training and inference~\cite{lepikhin2020scaling}. However, the inclusion of sparsity introduces a new challenge—\textit{load imbalance}—where a small number of experts are disproportionately activated while others remain underutilized. This imbalance not only reduces computational efficiency but also causes practical bottlenecks such as straggler-induced latency, idle compute resources, and increased energy consumption during distributed inference~\cite{he2025capacityawareinferencemitigatingstraggler,doucet2025harmoenyefficientmultigpuinference,Dhasade_2025}. The issue becomes increasingly severe with larger batch sizes or streaming workloads, leading to degraded throughput and unpredictable inference latency.

To address this, numerous works have explored balancing strategies primarily during the training phase. Early efforts integrate auxiliary loss functions to encourage balanced routing decisions~\cite{fedus2022switch,lepikhin2020gshard,zhou2022mixture} or impose hard expert capacity limits to prevent overload~\cite{lepikhin2020gshard}. Other studies employ stochastic or top-$k$ routing variants to diversify expert selection and reduce over-specialization~\cite{llama-moe-2023,zhou2022mixture,zuo2021taming,chen2023sparse,roller2021hash}. These methods improve balance within the training distribution but often fail to generalize at inference time, where routing dynamics differ substantially due to unseen inputs or batch-level aggregation effects. Beyond training-time balancing, SMoEs have been adopted in a wide range of applications, such as image segmentation and object recognition~\cite{eigen2015predicting,riquelme2021scaling,zhu2023exploring}, as well as large-scale LLMs like Mixtral~\cite{jiang2024mixtral} and DeepSeek-MoE~\cite{dai2024deepseekmoe}, demonstrating the scalability of expert-based modeling. However, efficient inference deployment remains an open challenge. 

Recent works have begun to address inference-time inefficiencies through both algorithmic and system-level innovations. MEGABLOCKS~\cite{gale2022megablocksefficientsparsetraining} improves sparse training throughput via block-sparse operations, while SwapMoE~\cite{kong2024swapmoeservingofftheshelfmoebased} dynamically loads and swaps experts at runtime to support edge deployment under constrained memory. Lina~\cite{li2023accelerating} introduces a distributed runtime scheduler that replaces overloaded experts with alternatives in the same top-$k$ candidate set, mitigating inference-time imbalance. Pre-gated MoE~\cite{hwang2024pregatedmoealgorithmsystemcodesign} introduces a pre-gating layer to reduce routing overhead and accelerate inference. For memory optimization, MoQE~\cite{kim2023mixturequantizedexpertsmoqe} combines expert quantization with sparse routing to further reduce activation cost. Similarly, MoE-Deployment~\cite{huang2023moedeploymentmitigatinginefficiencies} highlights the difficulty of balancing experts across GPUs during distributed inference, noting that naive scheduling can exacerbate computational skew.

More recently, several adaptive routing studies have moved beyond heuristic balancing rules, focusing instead on optimization-consistent and self-regularized strategies. Wang et al.~\cite{wang2024auxiliarylossfreeloadbalancingstrategy} propose an auxiliary-loss-free scheme that biases routing logits based on each expert’s recent activity, dynamically achieving smoother equilibrium. Omi et al.~\cite{omi2025loadbalancingmixtureexperts} introduce a similarity-preserving routing strategy that enforces consistent expert selection for semantically related tokens, improving both convergence and utilization. Despite these advances, most existing approaches either require retraining or architectural modification, leaving a notable gap in lightweight, training-free rebalancing methods—precisely the direction our work targets.

\paragraph{Improving Efficiency of LLM}
Enhancing the efficiency of large language models involves optimising hardware and developing algorithmic breakthroughs. In terms of leveraging hardware, using multiple GPUs or TPUs ~\cite{jouppi2017datacenter} significantly increased inference speed. Additionally, optimising model architecture can be an effective method. For example, quantization~\cite{gong2014compressing}~\cite{jacob2018quantization}~\cite{zhou2017incremental}~\cite{krishnamoorthi2018quantizing}, which reduces the numerical precision used in computation, effectively decreases model size and enhances inference speed without substantial accuracy loss. Recent works have further refined quantization granularity, demonstrating that intermediate-bit quantization (e.g., INT6)~\cite{zhang2025flexqefficientposttrainingint6} and post-training calibration frameworks~\cite{shi2025systematiccharacterizationllmquantization} can yield better accuracy–efficiency trade-offs for transformer-based LLMs. Additionally, algorithms such as pruning ~\cite{han2015learning}~\cite{molchanov2016pruning}~\cite{frankle2018lottery} and knowledge distillation~\cite{hinton2015distilling}~\cite{zagoruyko2016paying}~\cite{polino2018model} reduce the computational burden by removing irrelevant weights and teaching smaller models to imitate larger ones, respectively, which can still maintain or even improve the model's performance. Sparse training~\cite{mocanu2018scalable}~\cite{bellec2017deep}~\cite{mostafa2019parameter} was widely used to decrease the computational cost for the large models, which was performed by reducing the number of active neurons.
Recent advances have expanded these classical paradigms by coupling structural adaptation with data-driven compression, enabling LLMs to sustain generalization while achieving higher efficiency. For instance, LLM-Pruner~\cite{ma2023llmprunerstructuralpruninglarge} performs gradient-guided structured pruning that jointly removes attention heads and feed-forward blocks, followed by minimal fine-tuning to recover accuracy. Wanda~\cite{sun2024simpleeffectivepruningapproach} introduces activation-aware one-shot pruning that integrates both weight magnitude and activation statistics, allowing efficient compression without retraining.
Pruning itself is becoming more adaptive and theoretically grounded. Bypass Back-Propagation~\cite{gao2025bypassbackpropagationoptimizationbasedstructural} formulates pruning as a learnable mask-selection problem solved through policy optimization, avoiding explicit backpropagation through the full model. AlphaPruning~\cite{lu2024alphapruningusingheavytailedself} applies heavy-tailed regularization theory to derive optimal layerwise sparsity ratios, ensuring consistent gradient flow across depth. Dimension-Independent Structural Pruning (DISP-LLM)~\cite{gao2024displlmdimensionindependentstructuralpruning} further generalizes structured pruning by decoupling sparsity patterns across embedding dimensions, providing flexible control over compression granularity. Among these compression techniques, quantization has emerged as particularly synergistic with sparse architectures like MoE, as both aim to reduce computational overhead without retraining.
\paragraph{Model Quantization Method}
Model quantization has become an essential technique for improving the inference efficiency of large-scale neural networks, especially in resource-constrained settings. Early works such as vector quantization~\cite{gong2014compressingdeepconvolutionalnetworks}, integer-only inference~\cite{jacob2017quantizationtrainingneuralnetworks}, and incremental quantization~\cite{zhou2017incrementalnetworkquantizationlossless} laid the groundwork for compressing deep neural networks with minimal accuracy degradation. Subsequent methods like QAT (Quantization-Aware Training) and post-training quantization have been widely adopted across tasks. Notably, GPTQ~\cite{frantar2023gptqaccurateposttrainingquantization} and LLM.int8~\cite{dettmers2022llmint88bitmatrixmultiplication} introduced effective post-training quantization strategies tailored to transformer-based models. QLoRA~\cite{dettmers2023qloraefficientfinetuningquantized} advanced this further by combining low-rank adaptation with 4-bit quantization for fine-tuning, enabling low-resource deployments of large LLMs. MoQE~\cite{kim2023mixturequantizedexpertsmoqe} specifically addressed the synergy between MoE architectures and quantization. Additional studies have focused on mixed-precision training and hardware-aware techniques~\cite{yang2021bsqexploringbitlevelsparsity, Zadeh_2020}, demonstrating that careful quantization design can significantly reduce memory and computational cost while preserving model fidelity. These developments make quantization an ideal complement to sparsity-based frameworks such as MoE. Our proposed R\&Q method builds directly on this synergy by integrating quantization into inference-time expert rebalancing.

\section{Conclusion}
In conclusion, large language models with sparse mixture-of-experts (SMoE) architectures have shown empirical success across various tasks. This architecture allows SMoEs to scale up the number of experts without the need for a proportional increase in computational resources, offering an efficient way to improve performance on diverse tasks. However, despite these advantages, SMoE's sparse routing can lead to significant load imbalances among experts, some may become overloaded with too many tasks while others remain underutilized, causing efficiency issues during deployment. Our paper presents a plug-and-play strategy to address this load imbalance. We propose a low-cost method that identifies and replicates the heaviest expert using a lower-bit quantized version while also quantizing the least important expert to maintain the memory budget. We conducted thorough empirical evaluations to validate the effectiveness of this approach. The results indicate that our strategy successfully alleviates the load imbalance issue in SMoE architectures. Furthermore, the R\&Q strategy we used resulted in minimal loss of performance, making our method a practical and efficient improvement for the deployment of SMoE models in large-scale systems.

\section{Acknowledgment}
This research is partially supported by Meta SOTA Acceleration of Academic Collaboration (SAAC)

\bibliographystyle{assets/plainnat}
\bibliography{paper}

@article{jacobs1991adaptive,
  title={Adaptive mixtures of local experts},
  author={Jacobs, Robert A and Jordan, Michael I and Nowlan, Steven J and Hinton, Geoffrey E},
  journal={Neural computation},
  volume={3},
  number={1},
  pages={79--87},
  year={1991},
  publisher={MIT Press}
}

@article{shazeer2017outrageously,
  title={Outrageously large neural networks: The sparsely-gated mixture-of-experts layer},
  author={Shazeer, Noam and Mirhoseini, Azalia and Maziarz, Krzysztof and Davis, Andy and Le, Quoc and Hinton, Geoffrey and Dean, Jeff},
  journal={arXiv preprint arXiv:1701.06538},
  year={2017}
}

@article{lepikhin2020gshard,
  title={Gshard: Scaling giant models with conditional computation and automatic sharding},
  author={Lepikhin, Dmitry and Lee, HyoukJoong and Xu, Yuanzhong and Chen, Dehao and Firat, Orhan and Huang, Yanping and Krikun, Maxim and Shazeer, Noam and Chen, Zhifeng},
  journal={arXiv preprint arXiv:2006.16668},
  year={2020}
}

@article{fedus2022switch,
  title={Switch transformers: Scaling to trillion parameter models with simple and efficient sparsity},
  author={Fedus, William and Zoph, Barret and Shazeer, Noam},
  journal={Journal of Machine Learning Research},
  volume={23},
  number={120},
  pages={1--39},
  year={2022}
}

@article{zhou2022mixture,
  title={Mixture-of-experts with expert choice routing},
  author={Zhou, Yanqi and Lei, Tao and Liu, Hanxiao and Du, Nan and Huang, Yanping and Zhao, Vincent and Dai, Andrew M and Le, Quoc V and Laudon, James and others},
  journal={Advances in Neural Information Processing Systems},
  volume={35},
  pages={7103--7114},
  year={2022}
}

@inproceedings{eigen2015predicting,
  title={Predicting depth, surface normals and semantic labels with a common multi-scale convolutional architecture},
  author={Eigen, David and Fergus, Rob},
  booktitle={Proceedings of the IEEE international conference on computer vision},
  pages={2650--2658},
  year={2015}
}

@article{riquelme2021scaling,
  title={Scaling vision with sparse mixture of experts},
  author={Riquelme, Carlos and Puigcerver, Joan and Mustafa, Basil and Neumann, Maxim and Jenatton, Rodolphe and Susano Pinto, Andr{\'e} and Keysers, Daniel and Houlsby, Neil},
  journal={Advances in Neural Information Processing Systems},
  volume={34},
  pages={8583--8595},
  year={2021}
}

@article{zhu2023exploring,
  title={Exploring Sparse MoE in GANs for Text-conditioned Image Synthesis},
  author={Zhu, Jiapeng and Yang, Ceyuan and Zheng, Kecheng and Xu, Yinghao and Shi, Zifan and Shen, Yujun},
  journal={arXiv preprint arXiv:2309.03904},
  year={2023}
}

@article{lepikhin2020scaling,
  title={Scaling giant models with conditional computation and automatic sharding},
  author={Lepikhin, D and Lee, H and Xu, Y and Chen, D and Firat, O and Huang, Y and Krikun, M and Shazeer, N and Gshard, Z},
  journal={arXiv preprint arXiv:2006.16668},
  year={2020}
}

@inproceedings{zhao2023sparse,
  title={Sparse moe with language guided routing for multilingual machine translation},
  author={Zhao, Xinyu and Chen, Xuxi and Cheng, Yu and Chen, Tianlong},
  booktitle={The Twelfth International Conference on Learning Representations},
  year={2023}
}

@misc{llama-moe-2023,
  title={LLaMA-MoE: Building Mixture-of-Experts from LLaMA with Continual Pre-training},
  author={LLaMA-MoE Team},
  year={2023},
  month={Dec},
  url={https://github.com/pjlab-sys4nlp/llama-moe}
}

@article{zuo2021taming,
  title={Taming sparsely activated transformer with stochastic experts},
  author={Zuo, Simiao and Liu, Xiaodong and Jiao, Jian and Kim, Young Jin and Hassan, Hany and Zhang, Ruofei and Zhao, Tuo and Gao, Jianfeng},
  journal={arXiv preprint arXiv:2110.04260},
  year={2021}
}

@article{chen2023sparse,
  title={Sparse moe as the new dropout: Scaling dense and self-slimmable transformers},
  author={Chen, Tianlong and Zhang, Zhenyu and Jaiswal, Ajay and Liu, Shiwei and Wang, Zhangyang},
  journal={arXiv preprint arXiv:2303.01610},
  year={2023}
}

@article{roller2021hash,
  title={Hash layers for large sparse models},
  author={Roller, Stephen and Sukhbaatar, Sainbayar and Weston, Jason and others},
  journal={Advances in Neural Information Processing Systems},
  volume={34},
  pages={17555--17566},
  year={2021}
}

@inproceedings{jouppi2017datacenter,
  title={In-datacenter performance analysis of a tensor processing unit},
  author={Jouppi, Norman P and Young, Cliff and Patil, Nishant and Patterson, David and Agrawal, Gaurav and Bajwa, Raminder and Bates, Sarah and Bhatia, Suresh and Boden, Nan and Borchers, Al and others},
  booktitle={Proceedings of the 44th annual international symposium on computer architecture},
  pages={1--12},
  year={2017}
}

@article{gong2014compressing,
  title={Compressing deep convolutional networks using vector quantization},
  author={Gong, Yunchao and Liu, Liu and Yang, Ming and Bourdev, Lubomir},
  journal={arXiv preprint arXiv:1412.6115},
  year={2014}
}

@article{han2015learning,
  title={Learning both weights and connections for efficient neural network},
  author={Han, Song and Pool, Jeff and Tran, John and Dally, William},
  journal={Advances in neural information processing systems},
  volume={28},
  year={2015}
}

@article{hinton2015distilling,
  title={Distilling the knowledge in a neural network},
  author={Hinton, Geoffrey and Vinyals, Oriol and Dean, Jeff},
  journal={arXiv preprint arXiv:1503.02531},
  year={2015}
}

@article{mocanu2018scalable,
  title={Scalable training of artificial neural networks with adaptive sparse connectivity inspired by network science},
  author={Mocanu, Decebal Constantin and Mocanu, Elena and Stone, Peter and Nguyen, Phuong H and Gibescu, Madeleine and Liotta, Antonio},
  journal={Nature communications},
  volume={9},
  number={1},
  pages={2383},
  year={2018},
  publisher={Nature Publishing Group UK London}
}

@inproceedings{jacob2018quantization,
  title={Quantization and training of neural networks for efficient integer-arithmetic-only inference},
  author={Jacob, Benoit and Kligys, Skirmantas and Chen, Bo and Zhu, Menglong and Tang, Matthew and Howard, Andrew and Adam, Hartwig and Kalenichenko, Dmitry},
  booktitle={Proceedings of the IEEE conference on computer vision and pattern recognition},
  pages={2704--2713},
  year={2018}
}

@article{zhou2017incremental,
  title={Incremental network quantization: Towards lossless cnns with low-precision weights},
  author={Zhou, Aojun and Yao, Anbang and Guo, Yiwen and Xu, Lin and Chen, Yurong},
  journal={arXiv preprint arXiv:1702.03044},
  year={2017}
}

@article{krishnamoorthi2018quantizing,
  title={Quantizing deep convolutional networks for efficient inference: A whitepaper},
  author={Krishnamoorthi, Raghuraman},
  journal={arXiv preprint arXiv:1806.08342},
  year={2018}
}

@article{molchanov2016pruning,
  title={Pruning convolutional neural networks for resource efficient inference},
  author={Molchanov, Pavlo and Tyree, Stephen and Karras, Tero and Aila, Timo and Kautz, Jan},
  journal={arXiv preprint arXiv:1611.06440},
  year={2016}
}

@article{frankle2018lottery,
  title={The lottery ticket hypothesis: Finding sparse, trainable neural networks},
  author={Frankle, Jonathan and Carbin, Michael},
  journal={arXiv preprint arXiv:1803.03635},
  year={2018}
}

@article{zagoruyko2016paying,
  title={Paying more attention to attention: Improving the performance of convolutional neural networks via attention transfer},
  author={Zagoruyko, Sergey and Komodakis, Nikos},
  journal={arXiv preprint arXiv:1612.03928},
  year={2016}
}

@article{polino2018model,
  title={Model compression via distillation and quantization},
  author={Polino, Antonio and Pascanu, Razvan and Alistarh, Dan},
  journal={arXiv preprint arXiv:1802.05668},
  year={2018}
}

@article{bellec2017deep,
  title={Deep rewiring: Training very sparse deep networks},
  author={Bellec, Guillaume and Kappel, David and Maass, Wolfgang and Legenstein, Robert},
  journal={arXiv preprint arXiv:1711.05136},
  year={2017}
}

@inproceedings{mostafa2019parameter,
  title={Parameter efficient training of deep convolutional neural networks by dynamic sparse reparameterization},
  author={Mostafa, Hesham and Wang, Xin},
  booktitle={International Conference on Machine Learning},
  pages={4646--4655},
  year={2019},
  organization={PMLR}
}

@misc{jiang2024mixtralexperts,
      title={Mixtral of Experts}, 
      author={Albert Q. Jiang and Alexandre Sablayrolles and Antoine Roux and Arthur Mensch and Blanche Savary and Chris Bamford and Devendra Singh Chaplot and Diego de las Casas and Emma Bou Hanna and Florian Bressand and Gianna Lengyel and Guillaume Bour and Guillaume Lample and Lélio Renard Lavaud and Lucile Saulnier and Marie-Anne Lachaux and Pierre Stock and Sandeep Subramanian and Sophia Yang and Szymon Antoniak and Teven Le Scao and Théophile Gervet and Thibaut Lavril and Thomas Wang and Timothée Lacroix and William El Sayed},
      year={2024},
      eprint={2401.04088},
      archivePrefix={arXiv},
      primaryClass={cs.LG},
      url={https://arxiv.org/abs/2401.04088}, 
}

@misc{dai2024deepseekmoe,
  title = {DeepSeekMoE: Towards Ultimate Expert Specialization in Mixture-of-Experts Language Models},
  shorttitle = {DeepSeekMoE},
  author = {Dai, Damai and Deng, Chengqi and Zhao, Chenggang and Xu, R. X. and Gao, Huazuo and Chen, Deli and Li, Jiashi and Zeng, Wangding and Yu, Xingkai and Wu, Y. and Xie, Zhenda and Li, Y. K. and Huang, Panpan and Luo, Fuli and Ruan, Chong and Sui, Zhifang and Liang, Wenfeng},
  year = {2024},
  month = jan,
  number = {arXiv:2401.06066},
  eprint = {2401.06066},
  primaryclass = {cs},
  publisher = {arXiv},
  urldate = {2024-01-15},
  archiveprefix = {arxiv}
}

@article{wang2023mathcoder,
  title={Mathcoder: Seamless code integration in llms for enhanced mathematical reasoning},
  author={Wang, Ke and Ren, Houxing and Zhou, Aojun and Lu, Zimu and Luo, Sichun and Shi, Weikang and Zhang, Renrui and Song, Linqi and Zhan, Mingjie and Li, Hongsheng},
  journal={arXiv preprint arXiv:2310.03731},
  year={2023}
}

@article{yuan2023scaling,
  title={Scaling relationship on learning mathematical reasoning with large language models},
  author={Yuan, Zheng and Yuan, Hongyi and Li, Chengpeng and Dong, Guanting and Tan, Chuanqi and Zhou, Chang},
  journal={arXiv preprint arXiv:2308.01825},
  year={2023}
}

@article{imani2023mathprompter,
  title={Mathprompter: Mathematical reasoning using large language models},
  author={Imani, Shima and Du, Liang and Shrivastava, Harsh},
  journal={arXiv preprint arXiv:2303.05398},
  year={2023}
}

@article{huang2022towards,
  title={Towards reasoning in large language models: A survey},
  author={Huang, Jie and Chang, Kevin Chen-Chuan},
  journal={arXiv preprint arXiv:2212.10403},
  year={2022}
}

@article{jiang2024mixtral,
  title={Mixtral of experts},
  author={Jiang, Albert Q and Sablayrolles, Alexandre and Roux, Antoine and Mensch, Arthur and Savary, Blanche and Bamford, Chris and Chaplot, Devendra Singh and Casas, Diego de las and Hanna, Emma Bou and Bressand, Florian and others},
  journal={arXiv preprint arXiv:2401.04088},
  year={2024}
}

@misc{dettmers2022llmint88bitmatrixmultiplication,
      title={LLM.int8(): 8-bit Matrix Multiplication for Transformers at Scale}, 
      author={Tim Dettmers and Mike Lewis and Younes Belkada and Luke Zettlemoyer},
      year={2022},
      eprint={2208.07339},
      archivePrefix={arXiv},
      primaryClass={cs.LG},
      url={https://arxiv.org/abs/2208.07339}, 
}

@misc{gale2022megablocksefficientsparsetraining,
      title={MegaBlocks: Efficient Sparse Training with Mixture-of-Experts}, 
      author={Trevor Gale and Deepak Narayanan and Cliff Young and Matei Zaharia},
      year={2022},
      eprint={2211.15841},
      archivePrefix={arXiv},
      primaryClass={cs.LG},
      url={https://arxiv.org/abs/2211.15841}, 
}

@misc{kong2024swapmoeservingofftheshelfmoebased,
      title={SwapMoE: Serving Off-the-shelf MoE-based Large Language Models with Tunable Memory Budget}, 
      author={Rui Kong and Yuanchun Li and Qingtian Feng and Weijun Wang and Xiaozhou Ye and Ye Ouyang and Linghe Kong and Yunxin Liu},
      year={2024},
      eprint={2308.15030},
      archivePrefix={arXiv},
      primaryClass={cs.AI},
      url={https://arxiv.org/abs/2308.15030}, 
}

@misc{hwang2024pregatedmoealgorithmsystemcodesign,
      title={Pre-gated MoE: An Algorithm-System Co-Design for Fast and Scalable Mixture-of-Expert Inference}, 
      author={Ranggi Hwang and Jianyu Wei and Shijie Cao and Changho Hwang and Xiaohu Tang and Ting Cao and Mao Yang},
      year={2024},
      eprint={2308.12066},
      archivePrefix={arXiv},
      primaryClass={cs.LG},
      url={https://arxiv.org/abs/2308.12066}, 
}

@inproceedings{li2023accelerating,
  title={Accelerating distributed $\{$MoE$\}$ training and inference with lina},
  author={Li, Jiamin and Jiang, Yimin and Zhu, Yibo and Wang, Cong and Xu, Hong},
  booktitle={2023 USENIX Annual Technical Conference (USENIX ATC 23)},
  pages={945--959},
  year={2023}
}

@misc{kim2023mixturequantizedexpertsmoqe,
      title={Mixture of Quantized Experts (MoQE): Complementary Effect of Low-bit Quantization and Robustness}, 
      author={Young Jin Kim and Raffy Fahim and Hany Hassan Awadalla},
      year={2023},
      eprint={2310.02410},
      archivePrefix={arXiv},
      primaryClass={cs.LG},
      url={https://arxiv.org/abs/2310.02410}, 
}

@misc{huang2023moedeploymentmitigatinginefficiencies,
      title={Towards MoE Deployment: Mitigating Inefficiencies in Mixture-of-Expert (MoE) Inference}, 
      author={Haiyang Huang and Newsha Ardalani and Anna Sun and Liu Ke and Hsien-Hsin S. Lee and Anjali Sridhar and Shruti Bhosale and Carole-Jean Wu and Benjamin Lee},
      year={2023},
      eprint={2303.06182},
      archivePrefix={arXiv},
      primaryClass={cs.DC},
      url={https://arxiv.org/abs/2303.06182}, 
}

@misc{chowdhery2022palmscalinglanguagemodeling,
      title={PaLM: Scaling Language Modeling with Pathways}, 
      author={Aakanksha Chowdhery and Sharan Narang and Jacob Devlin and Maarten Bosma and Gaurav Mishra and Adam Roberts and Paul Barham and Hyung Won Chung and Charles Sutton and Sebastian Gehrmann and Parker Schuh and Kensen Shi and Sasha Tsvyashchenko and Joshua Maynez and Abhishek Rao and Parker Barnes and Yi Tay and Noam Shazeer and Vinodkumar Prabhakaran and Emily Reif and Nan Du and Ben Hutchinson and Reiner Pope and James Bradbury and Jacob Austin and Michael Isard and Guy Gur-Ari and Pengcheng Yin and Toju Duke and Anselm Levskaya and Sanjay Ghemawat and Sunipa Dev and Henryk Michalewski and Xavier Garcia and Vedant Misra and Kevin Robinson and Liam Fedus and Denny Zhou and Daphne Ippolito and David Luan and Hyeontaek Lim and Barret Zoph and Alexander Spiridonov and Ryan Sepassi and David Dohan and Shivani Agrawal and Mark Omernick and Andrew M. Dai and Thanumalayan Sankaranarayana Pillai and Marie Pellat and Aitor Lewkowycz and Erica Moreira and Rewon Child and Oleksandr Polozov and Katherine Lee and Zongwei Zhou and Xuezhi Wang and Brennan Saeta and Mark Diaz and Orhan Firat and Michele Catasta and Jason Wei and Kathy Meier-Hellstern and Douglas Eck and Jeff Dean and Slav Petrov and Noah Fiedel},
      year={2022},
      eprint={2204.02311},
      archivePrefix={arXiv},
      primaryClass={cs.CL},
      url={https://arxiv.org/abs/2204.02311}, 
}

@misc{sun2024simpleeffectivepruningapproach,
      title={A Simple and Effective Pruning Approach for Large Language Models}, 
      author={Mingjie Sun and Zhuang Liu and Anna Bair and J. Zico Kolter},
      year={2024},
      eprint={2306.11695},
      archivePrefix={arXiv},
      primaryClass={cs.CL},
      url={https://arxiv.org/abs/2306.11695}, 
}

@misc{gong2014compressingdeepconvolutionalnetworks,
      title={Compressing Deep Convolutional Networks using Vector Quantization}, 
      author={Yunchao Gong and Liu Liu and Ming Yang and Lubomir Bourdev},
      year={2014},
      eprint={1412.6115},
      archivePrefix={arXiv},
      primaryClass={cs.CV},
      url={https://arxiv.org/abs/1412.6115}, 
}

@misc{jacob2017quantizationtrainingneuralnetworks,
      title={Quantization and Training of Neural Networks for Efficient Integer-Arithmetic-Only Inference}, 
      author={Benoit Jacob and Skirmantas Kligys and Bo Chen and Menglong Zhu and Matthew Tang and Andrew Howard and Hartwig Adam and Dmitry Kalenichenko},
      year={2017},
      eprint={1712.05877},
      archivePrefix={arXiv},
      primaryClass={cs.LG},
      url={https://arxiv.org/abs/1712.05877}, 
}

@misc{zhou2017incrementalnetworkquantizationlossless,
      title={Incremental Network Quantization: Towards Lossless CNNs with Low-Precision Weights}, 
      author={Aojun Zhou and Anbang Yao and Yiwen Guo and Lin Xu and Yurong Chen},
      year={2017},
      eprint={1702.03044},
      archivePrefix={arXiv},
      primaryClass={cs.CV},
      url={https://arxiv.org/abs/1702.03044}, 
}

@misc{dettmers2023qloraefficientfinetuningquantized,
      title={QLoRA: Efficient Finetuning of Quantized LLMs}, 
      author={Tim Dettmers and Artidoro Pagnoni and Ari Holtzman and Luke Zettlemoyer},
      year={2023},
      eprint={2305.14314},
      archivePrefix={arXiv},
      primaryClass={cs.LG},
      url={https://arxiv.org/abs/2305.14314}, 
}

@misc{frantar2023gptqaccurateposttrainingquantization,
      title={GPTQ: Accurate Post-Training Quantization for Generative Pre-trained Transformers}, 
      author={Elias Frantar and Saleh Ashkboos and Torsten Hoefler and Dan Alistarh},
      year={2023},
      eprint={2210.17323},
      archivePrefix={arXiv},
      primaryClass={cs.LG},
      url={https://arxiv.org/abs/2210.17323}, 
}

@misc{yang2021bsqexploringbitlevelsparsity,
      title={BSQ: Exploring Bit-Level Sparsity for Mixed-Precision Neural Network Quantization}, 
      author={Huanrui Yang and Lin Duan and Yiran Chen and Hai Li},
      year={2021},
      eprint={2102.10462},
      archivePrefix={arXiv},
      primaryClass={cs.LG},
      url={https://arxiv.org/abs/2102.10462}, 
}

@inproceedings{Zadeh_2020,
   title={GOBO: Quantizing Attention-Based NLP Models for Low Latency and Energy Efficient Inference},
   url={http://dx.doi.org/10.1109/MICRO50266.2020.00071},
   DOI={10.1109/micro50266.2020.00071},
   booktitle={2020 53rd Annual IEEE/ACM International Symposium on Microarchitecture (MICRO)},
   publisher={IEEE},
   author={Zadeh, Ali Hadi and Edo, Isak and Awad, Omar Mohamed and Moshovos, Andreas},
   year={2020},
   month=oct, pages={811–824} }

@misc{hendrycks2021measuringmassivemultitasklanguage,
      title={Measuring Massive Multitask Language Understanding}, 
      author={Dan Hendrycks and Collin Burns and Steven Basart and Andy Zou and Mantas Mazeika and Dawn Song and Jacob Steinhardt},
      year={2021},
      eprint={2009.03300},
      archivePrefix={arXiv},
      primaryClass={cs.CY},
      url={https://arxiv.org/abs/2009.03300}, 
}

@misc{lin2022truthfulqameasuringmodelsmimic,
      title={TruthfulQA: Measuring How Models Mimic Human Falsehoods}, 
      author={Stephanie Lin and Jacob Hilton and Owain Evans},
      year={2022},
      eprint={2109.07958},
      archivePrefix={arXiv},
      primaryClass={cs.CL},
      url={https://arxiv.org/abs/2109.07958}, 
}

@misc{cobbe2021trainingverifierssolvemath,
      title={Training Verifiers to Solve Math Word Problems}, 
      author={Karl Cobbe and Vineet Kosaraju and Mohammad Bavarian and Mark Chen and Heewoo Jun and Lukasz Kaiser and Matthias Plappert and Jerry Tworek and Jacob Hilton and Reiichiro Nakano and Christopher Hesse and John Schulman},
      year={2021},
      eprint={2110.14168},
      archivePrefix={arXiv},
      primaryClass={cs.LG},
      url={https://arxiv.org/abs/2110.14168}, 
}

@misc{sakaguchi2019winograndeadversarialwinogradschema,
      title={WinoGrande: An Adversarial Winograd Schema Challenge at Scale}, 
      author={Keisuke Sakaguchi and Ronan Le Bras and Chandra Bhagavatula and Yejin Choi},
      year={2019},
      eprint={1907.10641},
      archivePrefix={arXiv},
      primaryClass={cs.CL},
      url={https://arxiv.org/abs/1907.10641}, 
}

@misc{zellers2019hellaswagmachinereallyfinish,
      title={HellaSwag: Can a Machine Really Finish Your Sentence?}, 
      author={Rowan Zellers and Ari Holtzman and Yonatan Bisk and Ali Farhadi and Yejin Choi},
      year={2019},
      eprint={1905.07830},
      archivePrefix={arXiv},
      primaryClass={cs.CL},
      url={https://arxiv.org/abs/1905.07830}, 
}

@misc{bisk2019piqareasoningphysicalcommonsense,
      title={PIQA: Reasoning about Physical Commonsense in Natural Language}, 
      author={Yonatan Bisk and Rowan Zellers and Ronan Le Bras and Jianfeng Gao and Yejin Choi},
      year={2019},
      eprint={1911.11641},
      archivePrefix={arXiv},
      primaryClass={cs.CL},
      url={https://arxiv.org/abs/1911.11641}, 
}

@inproceedings{yang-etal-2015-wikiqa,
    title = "{W}iki{QA}: A Challenge Dataset for Open-Domain Question Answering",
    author = "Yang, Yi  and
      Yih, Wen-tau  and
      Meek, Christopher",
    editor = "M{\`a}rquez, Llu{\'i}s  and
      Callison-Burch, Chris  and
      Su, Jian",
    booktitle = "Proceedings of the 2015 Conference on Empirical Methods in Natural Language Processing",
    month = sep,
    year = "2015",
    address = "Lisbon, Portugal",
    publisher = "Association for Computational Linguistics",
    url = "https://aclanthology.org/D15-1237/",
    doi = "10.18653/v1/D15-1237",
    pages = "2013--2018"
}

@misc{clark2022unifiedscalinglawsrouted,
      title={Unified Scaling Laws for Routed Language Models}, 
      author={Aidan Clark and Diego de las Casas and Aurelia Guy and Arthur Mensch and Michela Paganini and Jordan Hoffmann and Bogdan Damoc and Blake Hechtman and Trevor Cai and Sebastian Borgeaud and George van den Driessche and Eliza Rutherford and Tom Hennigan and Matthew Johnson and Katie Millican and Albin Cassirer and Chris Jones and Elena Buchatskaya and David Budden and Laurent Sifre and Simon Osindero and Oriol Vinyals and Jack Rae and Erich Elsen and Koray Kavukcuoglu and Karen Simonyan},
      year={2022},
      eprint={2202.01169},
      archivePrefix={arXiv},
      primaryClass={cs.CL},
      url={https://arxiv.org/abs/2202.01169}, 
}

@misc{kaplan2020scalinglawsneurallanguage,
      title={Scaling Laws for Neural Language Models}, 
      author={Jared Kaplan and Sam McCandlish and Tom Henighan and Tom B. Brown and Benjamin Chess and Rewon Child and Scott Gray and Alec Radford and Jeffrey Wu and Dario Amodei},
      year={2020},
      eprint={2001.08361},
      archivePrefix={arXiv},
      primaryClass={cs.LG},
      url={https://arxiv.org/abs/2001.08361}, 
}

@misc{wang2024auxiliarylossfreeloadbalancingstrategy,
      title={Auxiliary-Loss-Free Load Balancing Strategy for Mixture-of-Experts}, 
      author={Lean Wang and Huazuo Gao and Chenggang Zhao and Xu Sun and Damai Dai},
      year={2024},
      eprint={2408.15664},
      archivePrefix={arXiv},
      primaryClass={cs.LG},
      url={https://arxiv.org/abs/2408.15664}, 
}

@misc{omi2025loadbalancingmixtureexperts,
      title={Load Balancing Mixture of Experts with Similarity Preserving Routers}, 
      author={Nabil Omi and Siddhartha Sen and Ali Farhadi},
      year={2025},
      eprint={2506.14038},
      archivePrefix={arXiv},
      primaryClass={cs.LG},
      url={https://arxiv.org/abs/2506.14038}, 
}

@misc{zhang2025flexqefficientposttrainingint6,
      title={FlexQ: Efficient Post-training INT6 Quantization for LLM Serving via Algorithm-System Co-Design}, 
      author={Hao Zhang and Aining Jia and Weifeng Bu and Yushu Cai and Kai Sheng and Hao Chen and Xin He},
      year={2025},
      eprint={2508.04405},
      archivePrefix={arXiv},
      primaryClass={cs.LG},
      url={https://arxiv.org/abs/2508.04405}, 
}

@misc{shi2025systematiccharacterizationllmquantization,
      title={Systematic Characterization of LLM Quantization: A Performance, Energy, and Quality Perspective}, 
      author={Tianyao Shi and Yi Ding},
      year={2025},
      eprint={2508.16712},
      archivePrefix={arXiv},
      primaryClass={cs.PF},
      url={https://arxiv.org/abs/2508.16712}, 
}

@misc{ma2023llmprunerstructuralpruninglarge,
      title={LLM-Pruner: On the Structural Pruning of Large Language Models}, 
      author={Xinyin Ma and Gongfan Fang and Xinchao Wang},
      year={2023},
      eprint={2305.11627},
      archivePrefix={arXiv},
      primaryClass={cs.CL},
      url={https://arxiv.org/abs/2305.11627}, 
}

@misc{gao2025bypassbackpropagationoptimizationbasedstructural,
      title={Bypass Back-propagation: Optimization-based Structural Pruning for Large Language Models via Policy Gradient}, 
      author={Yuan Gao and Zujing Liu and Weizhong Zhang and Bo Du and Gui-Song Xia},
      year={2025},
      eprint={2406.10576},
      archivePrefix={arXiv},
      primaryClass={cs.LG},
      url={https://arxiv.org/abs/2406.10576}, 
}

@misc{lu2024alphapruningusingheavytailedself,
      title={AlphaPruning: Using Heavy-Tailed Self Regularization Theory for Improved Layer-wise Pruning of Large Language Models}, 
      author={Haiquan Lu and Yefan Zhou and Shiwei Liu and Zhangyang Wang and Michael W. Mahoney and Yaoqing Yang},
      year={2024},
      eprint={2410.10912},
      archivePrefix={arXiv},
      primaryClass={cs.LG},
      url={https://arxiv.org/abs/2410.10912}, 
}

@misc{gao2024displlmdimensionindependentstructuralpruning,
      title={DISP-LLM: Dimension-Independent Structural Pruning for Large Language Models}, 
      author={Shangqian Gao and Chi-Heng Lin and Ting Hua and Tang Zheng and Yilin Shen and Hongxia Jin and Yen-Chang Hsu},
      year={2024},
      eprint={2410.11988},
      archivePrefix={arXiv},
      primaryClass={cs.CL},
      url={https://arxiv.org/abs/2410.11988}, 
}

@misc{he2025capacityawareinferencemitigatingstraggler,
      title={Capacity-Aware Inference: Mitigating the Straggler Effect in Mixture of Experts}, 
      author={Shwai He and Weilin Cai and Jiayi Huang and Ang Li},
      year={2025},
      eprint={2503.05066},
      archivePrefix={arXiv},
      primaryClass={cs.LG},
      url={https://arxiv.org/abs/2503.05066}, 
}

@misc{doucet2025harmoenyefficientmultigpuinference,
      title={HarMoEny: Efficient Multi-GPU Inference of MoE Models}, 
      author={Zachary Doucet and Rishi Sharma and Martijn de Vos and Rafael Pires and Anne-Marie Kermarrec and Oana Balmau},
      year={2025},
      eprint={2506.12417},
      archivePrefix={arXiv},
      primaryClass={cs.DC},
      url={https://arxiv.org/abs/2506.12417}, 
}

@inproceedings{Dhasade_2025, series={EuroMLSys ’25},
   title={Practical Federated Learning without a Server},
   url={http://dx.doi.org/10.1145/3721146.3721938},
   DOI={10.1145/3721146.3721938},
   booktitle={Proceedings of the 5th Workshop on Machine Learning and Systems},
   publisher={ACM},
   author={Dhasade, Akash and Kermarrec, Anne-Marie and Lavoie, Erick and Pouwelse, Johan and Sharma, Rishi and de Vos, Martijn},
   year={2025},
   month=mar, pages={1–11},
   collection={EuroMLSys ’25} }

@misc{han2025gracemoegroupingreplicationlocalityaware,
      title={GRACE-MoE: Grouping and Replication with Locality-Aware Routing for Efficient Distributed MoE Inference}, 
      author={Yu Han and Lehan Pan and Jie Peng and Ziyang Tao and Wuyang Zhang and Yanyong Zhang},
      year={2025},
      eprint={2509.25041},
      archivePrefix={arXiv},
      primaryClass={cs.DC},
      url={https://arxiv.org/abs/2509.25041}, 
}

@misc{zoph2022stmoedesigningstabletransferable,
      title={ST-MoE: Designing Stable and Transferable Sparse Expert Models}, 
      author={Barret Zoph and Irwan Bello and Sameer Kumar and Nan Du and Yanping Huang and Jeff Dean and Noam Shazeer and William Fedus},
      year={2022},
      eprint={2202.08906},
      archivePrefix={arXiv},
      primaryClass={cs.CL},
      url={https://arxiv.org/abs/2202.08906}, 
}

@misc{lu2021codexgluemachinelearningbenchmark,
      title={CodeXGLUE: A Machine Learning Benchmark Dataset for Code Understanding and Generation}, 
      author={Shuai Lu and Daya Guo and Shuo Ren and Junjie Huang and Alexey Svyatkovskiy and Ambrosio Blanco and Colin Clement and Dawn Drain and Daxin Jiang and Duyu Tang and Ge Li and Lidong Zhou and Linjun Shou and Long Zhou and Michele Tufano and Ming Gong and Ming Zhou and Nan Duan and Neel Sundaresan and Shao Kun Deng and Shengyu Fu and Shujie Liu},
      year={2021},
      eprint={2102.04664},
      archivePrefix={arXiv},
      primaryClass={cs.SE},
      url={https://arxiv.org/abs/2102.04664}, 
}

@article{reddy-etal-2019-coqa,
    title = "{C}o{QA}: A Conversational Question Answering Challenge",
    author = "Reddy, Siva  and
      Chen, Danqi  and
      Manning, Christopher D.",
    editor = "Lee, Lillian  and
      Johnson, Mark  and
      Roark, Brian  and
      Nenkova, Ani",
    journal = "Transactions of the Association for Computational Linguistics",
    volume = "7",
    year = "2019",
    address = "Cambridge, MA",
    publisher = "MIT Press",
    url = "https://aclanthology.org/Q19-1016/",
    doi = "10.1162/tacl_a_00266",
    pages = "249--266"
}

@misc{merity2016pointer,
      title={Pointer Sentinel Mixture Models},
      author={Stephen Merity and Caiming Xiong and James Bradbury and Richard Socher},
      year={2016},
      eprint={1609.07843},
      archivePrefix={arXiv},
      primaryClass={cs.CL}
}

\end{document}